\documentclass{article}

\usepackage[final]{corl_2026}

\usepackage{graphicx}
\usepackage{booktabs}
\usepackage{rotating}   
\usepackage{makecell}   
\usepackage{amsmath}
\usepackage{xcolor}
\usepackage{microtype}
\usepackage{lipsum}     
\usepackage{adjustbox}
\usepackage{afterpage}
\usepackage{so101_preprint} 
\let\soOldTabular\tabular
\let\soOldEndTabular\endtabular
\renewenvironment{tabular}{\begin{adjustbox}{max width=\linewidth}\soOldTabular}{\soOldEndTabular\end{adjustbox}}

\makeatletter
\newcommand{\soInlineTableSetup}{\renewenvironment{table}[1][]{\def\@captype{table}}{}}
\makeatother
\graphicspath{{figures/}}

\newcommand{\todo}[1]{\textcolor{black}{[TODO: #1]}}

\title{SO-101 Bench: Accessible Spatial Manipulation}

\author{
  \textsf{\textbf{Truman Hickok}}\\
  Independent\\
  \texttt{hickoktruman@gmail.com}
}

\begin{document}
\begin{sofrontmatter}
\maketitle

{\small\sffamily
\href{https://so101-bench.github.io/}{\raisebox{-0.22em}{\includegraphics[height=1.3em]{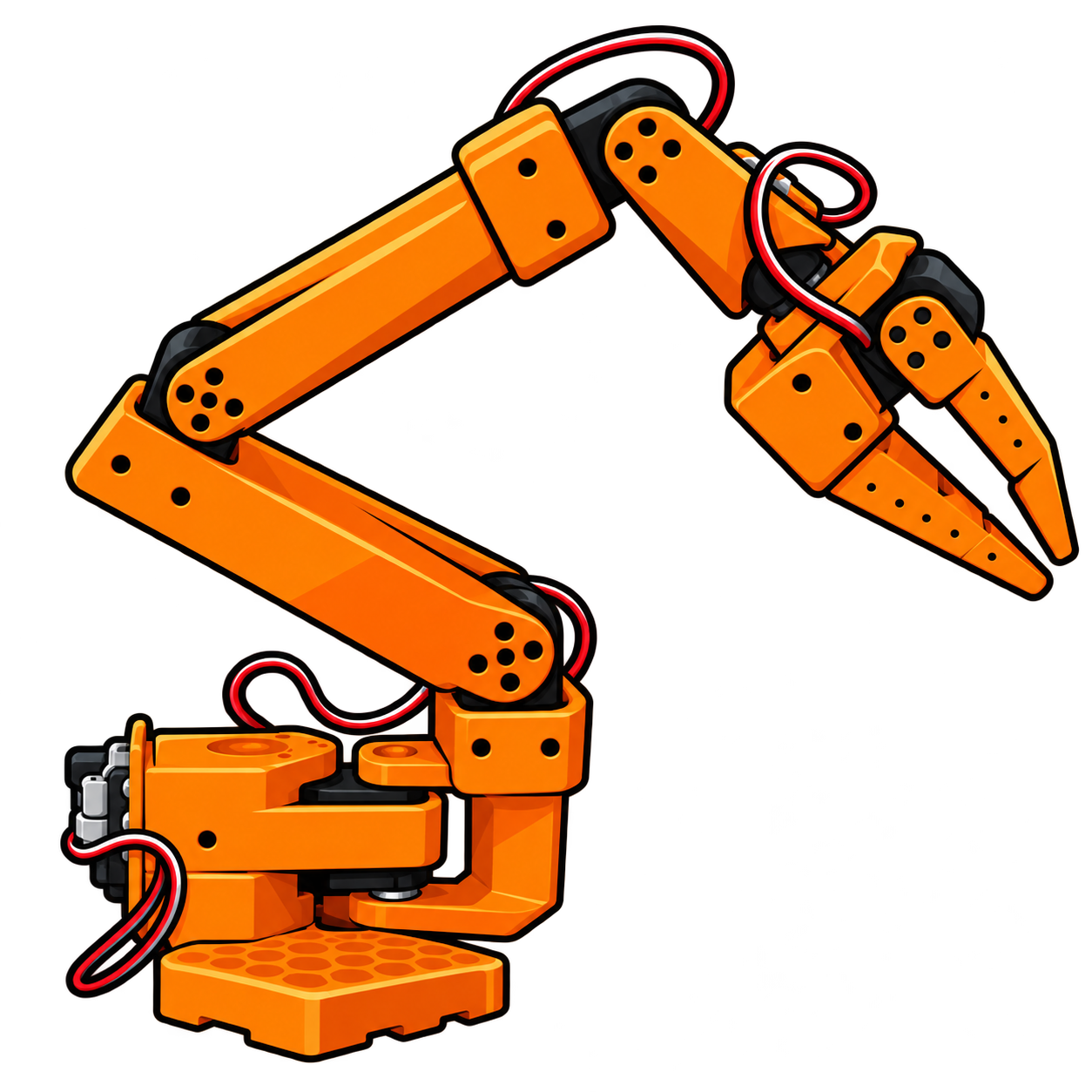}}\hspace{0.35em}Website}\quad\textperiodcentered\quad
\href{https://github.com/5hadytru/so101_bench}{\raisebox{-0.12em}{\includegraphics[height=1em]{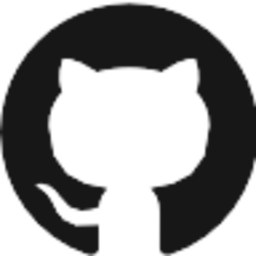}}\hspace{0.35em}GitHub}\quad\textperiodcentered\quad
\href{https://huggingface.co/5hadytru}{\raisebox{-0.18em}{\includegraphics[height=1.15em]{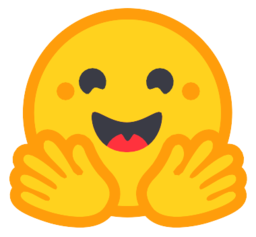}}\hspace{0.35em}Hugging Face}\par}
\vspace{0.7em}


\begin{abstract}
  Low-cost arms such as the SO-101 have become the most common entry point to robot learning, yet no released
  policy performs spatially constrained, language-conditioned tasks on them out of the box, and existing SO-101
  benchmarks run only on physical hardware with simple tasks and few held-out objects. We introduce SO-101 Bench, a
  simulated benchmark for language-conditioned spatial manipulation built on an Isaac Lab digital twin of a real
  SO-101 setup. It has four tasks scored as five families, 138 objects in seen, unseen-instance and unseen-class
  splits, 916 evaluation episodes with subsets for held-out appearances, rotate-to-fit planning and occlusion memory,
  and 3,629 simulated demonstrations on which every learned policy is fine-tuned under an equal compute budget. The
  twin ranks policies as the real robot does: five policies trained without real data, four VLAs and an LLM agent,
  keep their exact ranking when the same weights run on the physical arm in an unseen environment with unseen objects
  (Spearman $\rho=1.00$, MMRV $=0.000$), although simulation overestimates success. Across nine VLAs and world-action
  models, the best learned policies score near 61\%, with spatial placement, unseen object classes and rotate-to-fit
  planning remaining open. An LLM agent that writes code and hands grasping to a learned policy scores 83.2\%, and an
  ablation shows that this lead comes from the choice of harness and the choice of language model.
\end{abstract}

\keywords{Robot Manipulation, Benchmarks, Sim2Real}
\end{sofrontmatter}

\section{Introduction}
\label{sec:intro}

\suppressfloats[t] 
\begin{figure}[!t]
  \centering
  \includegraphics[width=\textwidth]{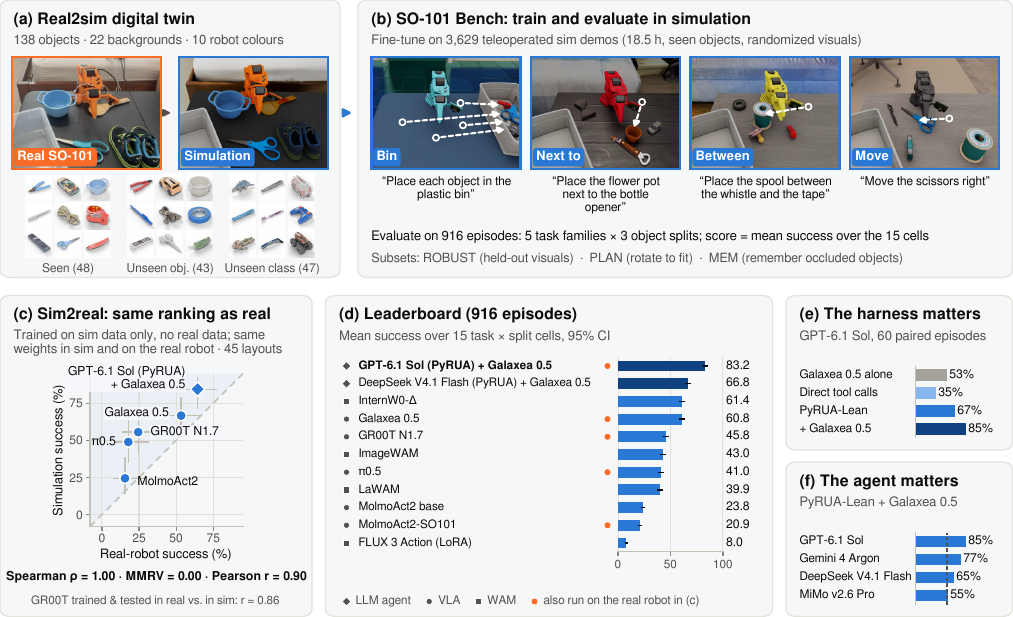}
  \caption{\textbf{SO-101 Bench at a glance.} (a) Real SO-101 and its digital twin. (b) The four tasks. (c) Sim2real:
  five policies trained without real data keep their ranking on the real robot. (d) Leaderboard. (e) Harness
  ablation. (f) The language model matters.}
  \label{fig:teaser}
\end{figure}

Low-cost arms have become the most common entry point to robot learning. The open-source
SO-101~\citep{hf2026so101,therobotstudio2024soarm100} is a 3D-printed arm driven by six serial-bus servos; a follower
arm costs about \$120--150 in parts~\citep{therobotstudio2024soarm100,selvaraj2026armnetbench}, against roughly
\$4,900 for one ALOHA arm~\citep{kawaharazuka2026mevion}. LeRobot~\citep{cadene2026lerobot} supports it from
teleoperation to deployment, and SO-family arms produce most of the community robot datasets on the Hugging Face
Hub.\footnote{559 of 976 (57\%) randomly sampled \texttt{LeRobot}-tagged datasets declare an SO-100/101
\texttt{robot\_type} in their metadata (our count, 26 October 2026).} Open models now train on these data: SmolVLA was
pretrained on 481 community datasets~\citep{shukor2025smolvla}, and MolmoAct2 has a variant post-trained on 38k
curated SO-100/101 episodes~\citep{fang2026molmoact2}. These models, however, are starting points for fine-tuning
on a user's own demonstrations: MolmoAct2-SO100/101 demonstrates out-of-the-box performance on basic tasks, and to our
knowledge no released policy performs spatially constrained, language-conditioned tasks on an SO-101 out of the box
in a new environment with new objects. Evaluation, too, has not kept pace: the SO-101 benchmarks released
this year run only on physical
hardware~\citep{yu2026so101failure,huang2026vlareplica,selvaraj2026armnetbench,tiensmith2026spabench}. Their tasks
include insertion, pouring, wiping and spatial grounding; Spa-Bench tests relational placement, referential descriptions,
ordinal position and counting, with familiar constituents in its primary evaluation and a separate novel-object
pilot~\citep{tiensmith2026spabench}. They are also costly:
real-robot studies typically afford tens of rollouts per condition~\citep{arkhangelskiy2026phail}, and results
depend on the particular build (servo offsets differ between arms~\citep{huang2026vlareplica}; one arm farm
reports a misaligned camera and objects that wore down during evaluation~\citep{selvaraj2026armnetbench}).

Simulation removes these costs, but a simulated benchmark is useful only if its rankings hold on hardware. Widely used
suites target other embodiments~\citep{liu2023libero,nasiriany2026robocasa365,chen2025robotwin2}, the one simulated
SO-101 task set we know of contains eight language-free cube and can tasks~\citep{almuzairee2026squint}, and the most
cited suite is saturated: a 0.54M-parameter policy reaches 95.1\% on LIBERO~\citep{sendai2026minerva}, where clean averages for three protocol-faithful
vision-language-action model (VLA) and world-action model (WAM) evaluations span only 1.36
points~\citep{wang2026industrialvlabench}. Real2sim
evaluation~\citep{li2024simpler,jain2025polaris,sedlacek2025realm,ranawaka2026simfoundry} shows that a carefully built
digital twin can predict real-world rankings, but the prediction is specific to the simulator, policies and setup on
which it was validated: in a five-policy study spanning 1,115 real rollouts across three simulators, SIMPLER
showed weak ranking agreement with the real robot~\citep{wang2026practical}. These studies also
mostly run in one direction, placing policies trained on real data inside a twin. As policies are increasingly trained
on simulated data alone~\citep{deshpande2026molmobot,yu2026oasis}, a benchmark should also show that its ranking
survives when the same sim-trained weights run on a real robot. Finally, it should measure what aggregate success
hides: a policy can complete a task without understanding it: on LIBERO, a probe that replaces instruction text with a task-ID embedding
nearly matches the best published results~\citep{jiang2026benchmarking}, and on a real SO-101, VLAs that manipulate
reliably can fail to adapt to OOD spatial relations~\citep{tiensmith2026spabench}.

We introduce SO-101 Bench, a simulated benchmark for language-conditioned spatial manipulation with a single SO-101
arm, built so that its rankings carry over to the real robot (Fig.~\ref{fig:teaser}). Its core is an Isaac
Lab~\citep{mittal2025isaaclab} digital twin of a real SO-101 setup with 138 household objects in three generalization
splits: seen (in the training dataset), unseen object of a seen class, and unseen class, with 37 unseen-class objects kept private. Four tasks
(place every object in a bin; place an object next to another; place it between two others; move it in a direction
until it reaches the boundary object) are scored automatically from simulator state. They require the policy to pick out the
named object among same-class distractors, grasp unfamiliar geometry and place it under spatial constraints. The
primary evaluation has 916 feasibility-checked episodes over 15 task-family $\times$ split cells, with subsets for
held-out appearances (ROBUST), rotating an object to fit a tight space (PLAN) and remembering occluded objects (MEM).
Every learned policy is fine-tuned on the same 3,629 simulated teleoperated demonstrations of seen objects under an
equal 36\,h x 4 x A100 compute budget.

Two studies test whether the twin can stand in for the real robot. Independently trained real-only and sim-only
GR00T~N1.6 policies find the same conditions hard (Pearson $r=0.86$ across the 15 cells) and fail in the same ways (Sec.~\ref{sec:correspondence}). Five policies that never saw real data from our setup,
four VLAs and an LLM agent, then run with identical weights in the twin and on the physical SO-101 in an environment
unseen in training, with objects from all three splits; SO-101 Bench policies work out of the box in unseen
real environments with unseen objects. Simulation ranks them exactly as the real robot does (Spearman $\rho=1.00$, MMRV $=0.000$, 95\% CI
$[0.000, 0.169]$), although it overestimates every policy's success by 9 to 31 percentage points
(Sec.~\ref{sec:sim2real}). On the full leaderboard of nine VLAs and WAMs, the best learned policies,
InternW0-$\Delta$~\citep{miao2026internw0} and Galaxea 0.5~\citep{liu2026g05}, are statistically tied near 61; spatial
placement and unseen classes are where they fail, and none exceeds 17\% on PLAN. The strongest entry is an LLM agent:
GPT-6.1 Sol~\citep{openai2026sol} writing code in the PyRUA-Lean harness~\citep{si2026pyrua} and handing grasps to
Galaxea 0.5 scores 83.2 and solves 75\% of PLAN. Its lead comes largely from the harness. On 60 paired episodes, Sol
succeeds on 35\% (21/60) with direct tool calls, 67\% (40/60) with PyRUA-Lean and 85\% (51/60) with the Galaxea 0.5
hand-off, against 53\% (32/60) for Galaxea 0.5 alone, and the lead holds on the real robot, 64\% (29/45) against 53\%
(24/45).

Our contributions are:
\begin{itemize}
  \item \textbf{SO-101 Bench}, a digital-twin benchmark for language-conditioned spatial manipulation on a low-cost
  arm, with instance- and class-level object splits, ROBUST, PLAN and MEM subsets and automatic scoring, released with
  the \texttt{so101\_bench\_sim} (3,629 simulated episodes) and \texttt{so101\_bench\_real} ($\sim$4,400 real
  episodes) datasets.
  \item \textbf{Evidence that the twin ranks policies as the real robot does}: condition- and failure-level
  correspondence between independently trained real-only and sim-only policies, and an exact sim2real ranking of five
  policies, including an LLM agent, run with identical weights and no real data from our setup; these policies
  perform language-conditioned tasks out of the box in an unseen real environment with unseen objects.
  \item \textbf{A leaderboard of nine VLAs and WAMs and two LLM agents}, with every learned policy fine-tuned on the
  same data and compute, showing that spatial placement, unseen object classes and rotate-to-fit planning remain open.
  \item \textbf{Harness and model ablations} showing that, in our setting, a code-as-policy harness that hands
  grasping to a learned policy outperforms either on their own; Direct action generation by the agent proves expensive and ineffective, and 6.1 Sol compares favorably to other agents.
\end{itemize}


\section{Related Work}
\label{sec:related}

\paragraph{Low-cost hardware and SO-101 benchmarks.} Cheap hardware has repeatedly widened who can do robot learning.
ALOHA showed that a low-cost bimanual rig can learn fine manipulation~\citep{zhao2023aloha}, and 2026 platforms push
prices further, from a \$1,688 bimanual mobile manipulator~\citep{li2026noria3} to the SO-101, at about a fortieth of
the price of an ALOHA arm~\citep{kawaharazuka2026mevion}. LeRobot~\citep{cadene2026lerobot} and community data have
made the SO-100/101 a training platform for open VLAs~\citep{shukor2025smolvla,fang2026molmoact2}. Four 2026 benchmarks evaluate VLAs on
physical SO-101 arms. \citet{yu2026so101failure} analyse failure and recovery on four tasks, VLA-REPLICA is a
sub-\$1,100 kit reproduced across sites~\citep{huang2026vlareplica}, ArmnetBench scores 2,518 rollouts of seven
policies on an arm farm~\citep{selvaraj2026armnetbench}, and Spa-Bench probes recombinations of familiar objects, spatial concepts and actions
in its primary evaluation, with a separate novel-object pilot~\citep{tiensmith2026spabench}. All four use real
robot demonstrations for task-specific adaptation and evaluate on physical hardware. In simulation, the SO-101 Task
Set contains eight language-free tasks for visual reinforcement learning with sim2real
transfer~\citep{almuzairee2026squint}, and LeIsaac provides SO-101 teleoperation and data-collection tooling in Isaac
Lab~\citep{lightwheel2025leisaac}. To our knowledge, SO-101 Bench is the first language-conditioned SO-101 benchmark
with held-out object splits whose simulated policy ranking is validated on the real arm. It also represents one of, if not the most, challenging task suites 
handled by an SO-101 up to this point.

\paragraph{Real2sim evaluation and sim2real transfer.} Whether simulated evaluation predicts real performance was
studied early for navigation~\citep{kadian2020sim2real}; for manipulation, SIMPLER~\citep{li2024simpler} showed that
an aligned simulator can rank real-world policies and introduced the MMRV metric. Twins built from scans,
neural reconstructions or generated assets now report strong score or ranking
agreement within their validated settings~\citep{jain2025polaris,sedlacek2025realm,yang2026robolab,ruan2026x2real,xiao2026h2rbench,wang2026reveal};
SimFoundry reaches Pearson 0.911 and MMRV 0.018 over five policy architectures~\citep{ranawaka2026simfoundry}. Two
2026 results temper this. Validity does not carry across setups: re-tested against new real rollouts, SIMPLER's
agreement with the real-robot ranking is weak~\citep{wang2026practical}. Reconstruction quality also matters: an authored object-and-scene pipeline achieved sim--real correlation of 0.90
versus 0.51 for a default reconstruction in a ten-cell study~\citep{verma2026measuring}, consistent with our tuning of each reconstructed object against its
physical counterpart (Appendix~\ref{app:objects}). Learned world models are an alternative
evaluator~\citep{geminirobotics2025veo}, but released models often predict success on verified
failures~\citep{xu2026worldmodelsdream}. In parallel, policies trained only on simulated data now transfer to real
robots: MolmoB0T reaches 79.2\% on real pick-and-place without real demonstrations~\citep{deshpande2026molmobot},
OASIS pairs image-to-3D assets with teleoperation inside simulation~\citep{yu2026oasis}, and LEGS and Splat2Act fine-tune
pretrained VLAs using only synthetic robot demonstrations~\citep{kim2026legs,ye2026splat2act}. We deploy five sim-trained policies, including an LLM agent, with identical
weights and test whether their ranking survives.

\paragraph{LLM agents for manipulation.} Language models have long planned over learned skills~\citep{ahn2022saycan}
or written programs that compose perception and control~\citep{liang2023codeaspolicies,huang2023voxposer}, and
hierarchical systems pair high-level language reasoning with low-level action
generation~\citep{shi2025hirobot,pi2025pi05}. In 2026 such agents became competitive with end-to-end policies: coding
agents are now benchmarked on manipulation directly~\citep{fu2026capx,lyu2026codeactionbench,chu2026openrua}; GPT-6
Astra with direct end-effector tool calls ranked above all forty public-policy entries in its September 2026
RoboDojo comparison~\citep{zhang2026unexpectedpolicy,chen2026robodojo}; RoboICL reports higher aggregate progress
than its strongest listed baseline on a thirty-task subset~\citep{liu2026roboicl}, and PyRUA-Lean exceeds the best
public entries in its reported leaderboard snapshot~\citep{si2026pyrua}. A parallel line treats the VLA as a tool: Harness VLA exposes
a frozen VLA as a retryable contact-rich primitive~\citep{zhang2026harnessvla}, VoLo steers one as an interruptible
tool~\citep{chen2026volo}, Gemini Robotics ER~2 hands motor execution to any VLA~\citep{hansen2026geminiroboticser2},
and a systematic study of such hierarchies finds that model choice and interface jointly shape
performance~\citep{hu2026hivla}. Harness effects are large. With the same planner and primitives, one study reports that writing code instead
of calling tools raises success from 63.1\% to 71.7\%~\citep{si2026pyrua}; a better harness lets a weaker model beat a
stronger one~\citep{li2026roboharnessk1}; and outside robotics the scaffold alone moves accuracy by up to 28 points
within one model~\citep{starace2026scaffoldgaia}. These studies usually compare agents with published entries trained
under each entry's own protocol. Simulated agent evaluations, ours included, commonly pause the simulator while the
model computes, which hides its real-time cost~\citep{sensenova2026astra}, and frontier models controlling physical
arms directly are not yet precise or efficient enough to finish tasks within a practical action
budget~\citep{leet2026bitterlesson}. SO-101 Bench compares agents and learned policies on the same episodes after
training every learned policy on the same data and compute, ablates harness and model separately, and runs the agent
on the real robot.

\paragraph{Manipulation benchmarks and evaluation practice.} Simulated benchmarks for generalist
policies~\citep{liu2023libero,nasiriany2026robocasa365,chen2025robotwin2,chen2026robodojo,zhang2025vlaarena} face
saturation and train--test overlap. LIBERO instruction conditioning largely selects among memorized
tasks~\citep{sendai2026minerva}; policies above 90\% there fall below 30\%, or to zero, under modest
perturbations~\citep{fei2025liberoplus,zhou2025liberopro}; and an audit of five suites finds shortcut solvability,
state-of-the-art gains not established as statistically significant, creeping overfitting and dependence on the training-data
source~\citep{jiang2026benchmarking}. Responses include generalization
suites~\citep{wang2026liberox,zhang2025vlaarena}, capability suites for memory of occluded
objects~\citep{dai2026robomme} and spatial grounding~\citep{tiensmith2026spabench}. Real-world evaluation has moved to
distributed pairwise comparison~\citep{atreya2025roboarena}, autonomous evaluation cells~\citep{zhou2025autoeval}, and
statistical methods that separate policies with few trials~\citep{arkhangelskiy2026phail,badithela2026anyrank}. SO-101
Bench adopts several of these lessons: it separates unseen instances from unseen classes and keeps part of the latter
private, fine-tunes every policy on the same seen-object data under one compute budget, selects checkpoints on a
separate validation set, and reports confidence intervals over 916 episodes. It also scores two spatial skills that,
to our knowledge, few benchmarks test in closed loop: placing an object between two referents, and rotating a grasped
object to fit (PLAN).


\clearpage
\begin{figure}[!t]
  \centering
  \includegraphics[width=\textwidth]{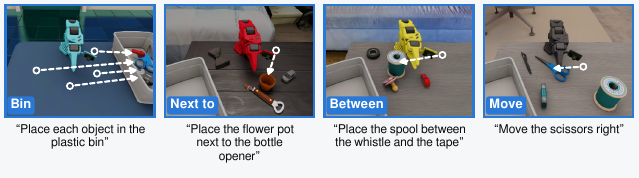}
  \caption{\textbf{The four SO-101 Bench tasks.} Bin, next-to, between and directional move, with the
  target paths overlaid on successful simulation episodes. The task panels repeat Fig.~\ref{fig:teaser}b.}
  \label{fig:tasks}
\end{figure}

\section{SO-101 Bench}
\label{sec:benchmark}

SO-101 Bench evaluates language-conditioned tabletop manipulation with a single SO-101 arm~\citep{hf2026so101}. It is designed to
test whether a policy can turn semantic understanding into precise physical execution: identifying the right
object among distractors, grasping objects of unfamiliar geometry, and placing them under (often precise) spatial constraints.
Every component below is shared by all evaluated policies.

\subsection{Tasks}
\label{sec:tasks}

The benchmark comprises four language-conditioned tasks (Fig.~\ref{fig:tasks}), scored as five task families.
\begin{itemize}
  \item \textbf{Place each object in the plastic bin.} The pure grasping task: the policy must identify, grasp,
  transport and deposit every object on the table. The bin moves between episodes. We evaluate a 1-object and a
  4-object condition as separate families.
  \item \textbf{Place [object 1] next to [object 2].} The policy must place the target close to a referent without
  disturbing other objects, adding semantic grounding and basic spatial placement.
  \item \textbf{Place [object 1] between [object 2] and [object 3].} The target must end centred between two
  referents, which makes the task sensitive to object extent, orientation and fine placement; it is the most demanding task.
  \item \textbf{Move [object 1] [direction].} The policy must push or carry the target along a straight path in a
  commanded direction until it reaches the boundary formed by the next object in its way.
\end{itemize}
The bin task stresses grounding, grasp strategy and grasp precision; the three spatial tasks additionally require
relational grounding, spatial reasoning, fine-grained placement and robustness to partial occlusion. To keep
language grounding nontrivial, distractors of the same class or colour are included so that class identity alone
is insufficient, and colour words are omitted from instructions unless necessary (i.e when two objects of the same class are on the table).

\clearpage
\begin{figure}[t]
  \centering
  \includegraphics[width=\textwidth]{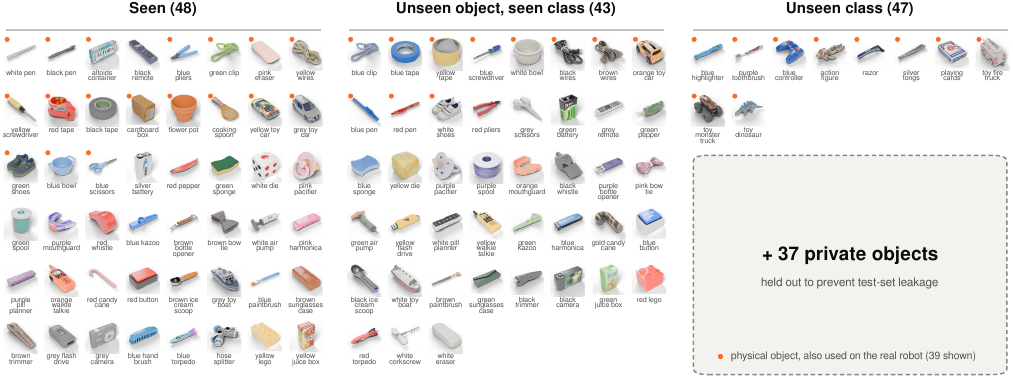}
  \caption{\textbf{Simulation objects} by generalization split. Orange dots mark physical objects that are also
  used in the real-robot evaluation. The 37 private unseen-class objects are held out
  to prevent test-set leakage and are not shown.}
  \label{fig:objects}
\end{figure}

\subsection{Objects and Generalization Splits}
\label{sec:objects}

The benchmark uses 138 household objects in three generalization splits (Fig.~\ref{fig:objects}):
\textbf{seen} (48 objects that appear in training), \textbf{unseen object, seen class} (43 objects whose class,
e.g.\ bowl or pen, appears in training but whose instance does not), and \textbf{unseen class} (47 objects of
entirely novel classes). The splits separate transfer across \emph{instances} from transfer across
\emph{categories}. Objects were chosen to maximize diversity of geometry and of domain (e.g.\ office, kitchen,
sports, toys), with priority on irregular shapes. Every object was verified through teleoperation to be
graspable by the SO-101, so that failures reflect the policy rather than the hardware. Of the unseen-class
objects, 37 are kept private to prevent test-set leakage.

All objects were created using the image-to-3D tools in 3D AI Studio~\citep{3daistudio2026}. The $\sim$60 physical objects were reconstructed
from photographs of the real items and their simulated appearance and physical properties were tuned to match
them; the remaining objects were generated from internet or AI-generated images and passed through the same
pipeline (Appendix~\ref{app:objects}). All simulated objects are rigid bodies.

\subsection{Digital Twin}
\label{sec:twin}

All evaluation runs in a digital twin of the original real-world setup, built in Isaac Lab~\citep{mittal2025isaaclab} from a scan of the room
and the object assets of Sec.~\ref{sec:objects}. Physics runs at 240\,Hz and the policy acts at 30\,Hz. Policies observe an overhead
camera and a wrist camera, both at $640\times480$, and the arm's joint positions. The twin includes 22 background
environments: 18 are used in training and in the varied evaluation episodes, and 4 are held out
(Appendix~\ref{app:backgrounds}). The robot is rendered in one of 6 training colours or one of 4 held-out colours.
The real arm's joint calibration is aligned to the simulated kinematics so that the same policy weights can run in
both domains (Sec.~\ref{sec:sim2real}).

\subsection{Scoring}
\label{sec:scoring}

Episodes are scored automatically from simulator state. Positions are measured in the tabletop plane from where
objects rest at the start of the episode. A success condition must hold continuously for 5\,s; a failure condition
ends the episode once it has held for 5\,s, so briefly undoing a mistake is allowed. The robot is considered to
touch an object when any link presses on it with more than 0.05\,N. Episodes time out after 30\,s (75\,s for
4-object bins).
\begin{itemize}
  \item \textbf{Bin:} every object ends with its reference point (the centre of its top-down bounding box) inside
  the bin's footprint and untouched by the robot. Fails if the bin moves more than 0.5\,in.
  \item \textbf{Next to:} the target's top-down outline ends within 2\,in of the referent's outline.
  \item \textbf{Between:} the target's reference point ends within 1.75\,in of the line through the two referents'
  reference points, and closer to each referent than the referents are to each other.
  \item \textbf{Move:} with an object ahead of the target, the target's leading edge ends within 2\,in of that
  object's facing edge; otherwise the target advances at least 2\,in. The target must stay within 2\,in of the
  straight path, and the boundary object must not move more than 0.5\,in along the direction.
\end{itemize}
In all three spatial tasks the target must move at least 1\,in and must not be touched by the robot at the end,
and the episode fails if any non-target object (referents included) moves more than 1\,in, or if the target or a
held object stays in contact with another object for more than 5\,s.

\subsection{Training Data}
\label{sec:data}

All learned policies are fine-tuned on \texttt{so101\_bench\_sim}: 3,629 teleoperated episodes (18.5\,h at 30\,fps)
recorded in the digital twin using seen objects only. It contains 1,892 bin episodes (including bin instructions
that name a single target among other objects), 1,150 next-to and between episodes, and 587 move episodes; 3,004
episodes have four objects on the table. Most episodes are rendered with randomized lighting, camera intrinsics and
extrinsics, background and robot colour; the canonical setup remains in the training distribution. A variant,
\texttt{so101\_bench\_sim\_WM}, adds the episode's initial overhead and wrist frames for policies that accept them.
We also release \texttt{so101\_bench\_real}, approximately 4,400 real-robot teleoperated episodes ($\sim$20\,h)
across approximately 35 objects.

\subsection{Primary Evaluation}
\label{sec:primary_eval}

\begin{figure}[t]
  \centering
  \includegraphics[width=\textwidth]{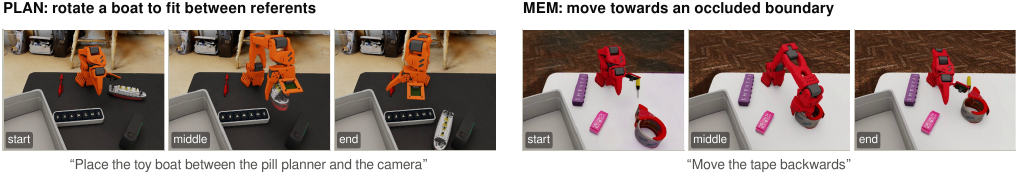}
  \caption{\textbf{PLAN and MEM examples} (start, middle and end of successful primary-evaluation episodes).
  Left: GPT-6.1 Sol with PyRUA-Lean and Galaxea 0.5 rotates the toy boat to fit between the pill planner
  and the camera (episode 529), which all nine learned policies fail.
  Right: ImageWAM moves the tape backwards (episode 300), retaining the boundary object
  position while it is occluded during the move; four of the nine learned policies solve this episode.
  All objects and backgrounds shown are public.}
  \label{fig:plan_mem}
\end{figure}

The primary evaluation contains 916 scored episodes spanning the 15 task-family $\times$ split cells, with every
object appearing in each family it supports. Feasibility (spacing between objects) is checked for every layout so
that a poor arrangement cannot make success impossible. About half the episodes use the canonical appearance and
half vary the background and robot colour. We report the \textbf{benchmark score}, the mean success rate over the
15 cells, with a 95\% bootstrap confidence interval that resamples episodes within cells. Three overlapping subsets
probe specific capabilities: \textbf{ROBUST} (178 episodes with held-out backgrounds and robot colours),
\textbf{PLAN} (48 episodes in which the grasped object must be rotated to fit into a tight space), and
\textbf{MEM} (51 episodes in which the policy must remember the position and 3D footprint of an object that becomes
occluded). Figure~\ref{fig:plan_mem} shows a successful example of each.

\subsection{Fine-Tuning Protocol}
\label{sec:protocol}

Every learned policy is fine-tuned with an equal compute budget of 36\,h on 4$\times$A100-80GB GPUs, following its
authors' recommended recipe where one exists (Appendix~\ref{app:s3}). Checkpoints are evaluated on a held-out
39-episode validation set, and the best checkpoint is used for the primary evaluation. Three policies (GR00T~N1.7~\citep{nvidia2025gr00tn1,nvidia2026gr00tn17},
$\pi_{0.5}$~\citep{pi2025pi05} and MolmoAct2~\citep{fang2026molmoact2}) additionally receive the episode's initial frames at each timestep.


\section{Real-Only vs.\ Sim-Only Policy Correspondence}
\label{sec:correspondence}

\begin{figure}[t]
  \centering
  \includegraphics[width=\textwidth]{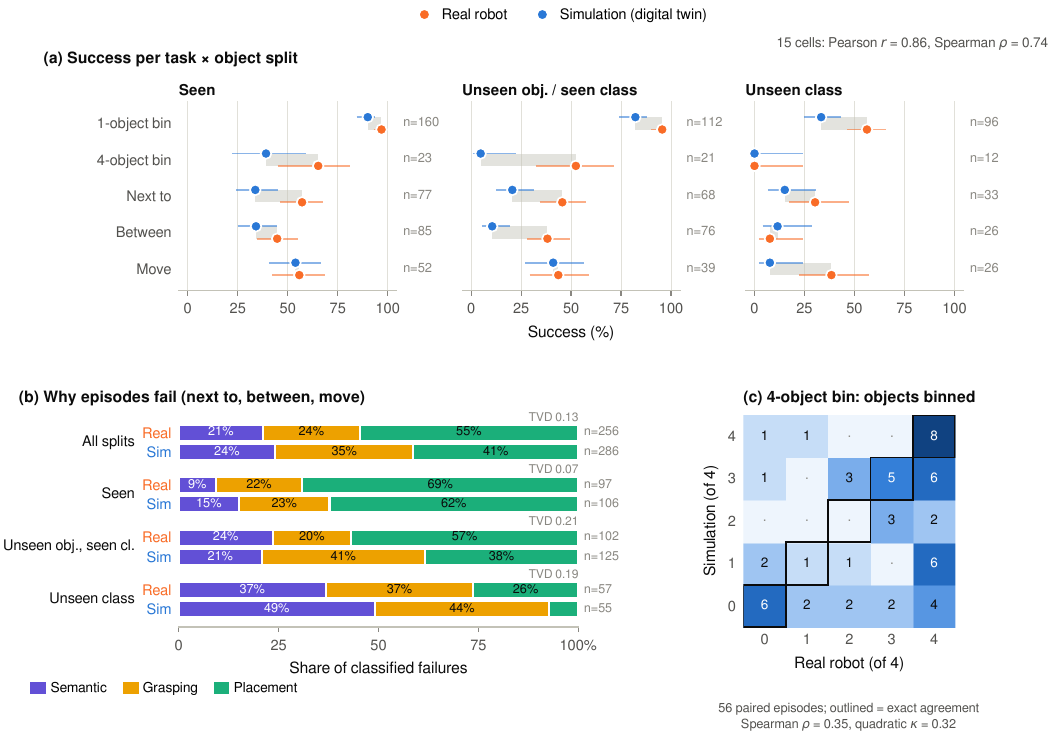}
  \par\vspace{5pt}
  \includegraphics[width=0.72\textwidth]{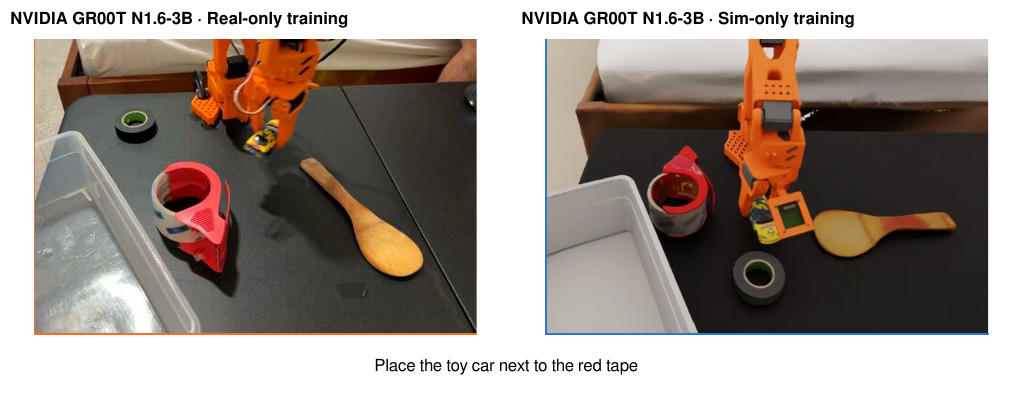}
  \caption{\textbf{Sim--real correspondence of independently trained GR00T N1.6 policies.} (a) Success of a
  real-only policy on the physical SO-101 and a sim-only policy in the digital twin, per task and object split
  (95\% Wilson intervals). (b) Composition of hand-annotated failures (next to, between, move): both domains shift
  from placement to semantic and grasping failures as object novelty increases. (c) Objects binned in 4-object
  bin episodes, real vs.\ simulation (56 paired episodes; outlined cells are exact agreement).
  Below: overhead views of independently trained NVIDIA GR00T N1.6-3B policies executing the same
  instruction with the same objects (real trial 52, matched task row 466). Poses and frame times differ.}
  \label{fig:s1}
\end{figure}

Before measuring sim2real transfer (Sec.~\ref{sec:sim2real}), we ask a coarser question:
does the digital twin reproduce the \emph{difficulty structure} of the real benchmark? If a policy trained in
simulation and evaluated in the twin finds the same tasks, splits and failure modes hard as a policy trained and
evaluated in the real world, the twin is a useful proxy for real-world evaluation even before any weights cross
domains.

\paragraph{Setup.} We fine-tune two GR00T~N1.6-3B policies~\citep{nvidia2025gr00tn1,nvidia2025gr00tn16} independently: a \emph{real-only} policy on real
teleoperated demonstrations (\texttt{so101\_bench\_real}), evaluated on the physical SO-101, and a \emph{sim-only} policy on demonstrations
collected in the digital twin (a subset of \texttt{so101\_bench\_sim} containing 3D scanned objects from \texttt{so101\_bench\_real}, with ~50\% of the data volume of \texttt{so101\_bench\_real}), evaluated in simulation. 
We compare them on 906 episodes matched across domains by task, instruction, ordered object set and object split.
Object poses are not identical across domains, so individual episodes are not exact replicas; the primary
comparison is therefore at the level of the 15 task $\times$ split conditions and of failure composition, with
episode-level agreement reported as a secondary measure. Real-world outcomes and failure types were labelled by
hand; simulated failures were labelled the same way and both were mapped to three categories, \emph{semantic}
(wrong object or referent(s)), \emph{grasping} (grasp, lift or release failed) and \emph{placement} (correct object
handled, imprecise/incorrect placement).

\paragraph{Difficulty is preserved across conditions.} The sim-only policy succeeds on 45\% (409/906) of episodes
and the real-only policy on 61\% (552/906), a gap of $-15.8$\,pp (95\% CI $[-19.9, -11.8]$). Despite this offset,
the two systems order the 15 conditions similarly: Pearson $r=0.86$ $[0.71, 0.91]$ and Spearman $\rho=0.74$
$[0.52, 0.91]$ across cells (Fig.~\ref{fig:s1}a). In both domains success is highest on the 1-object bin and
seen objects and falls with task difficulty and object novelty, with the 4-object bin and the between task
hardest on unseen classes.

\paragraph{Failure modes are preserved.} The composition of failures on the spatial tasks follows the same trend
in both domains (Fig.~\ref{fig:s1}b). On seen objects most failures are placement errors (69\% real, 62\% sim;
total variation distance 0.07); on unseen classes most are semantic or grasping errors (74\% real, 93\% sim).
Across all splits the two failure mixes differ by a total variation distance of 0.13. On the 4-object bin, the
real-only policy bins 67\% of objects and the sim-only policy 47\%; per-episode progress is weakly
correlated across domains, likely due to the small size of the sim dataset for this experiments (Spearman $\rho=0.35$; Fig.~\ref{fig:s1}c).

\paragraph{Episode-level agreement is moderate.} Paired episodes have the same outcome 66\% of the time
(MCC 0.36), and where both failed the failure type matches in 47\% of cases (Cohen's $\kappa=0.17$). This is
expected given differeing training dataset scales, initial poses and stochastic policies, and is why we emphasize condition-level
correspondence. Full results are in Appendix~\ref{app:s1}.

Overall, because the two policies differ in training data as well as in evaluation domain,
this experiment cannot attribute the absolute gap to the simulator; the lower sim success reflects both policies
and domains together. What it does show is that the twin reproduces which conditions are hard and why policies
fail there. Section~\ref{sec:sim2real} isolates the simulator by running identical weights (with sim training only) in both domains.


\section{Sim2Real with No Real Data}
\label{sec:sim2real}

\begin{figure}[t]
  \centering
  \includegraphics[width=\textwidth]{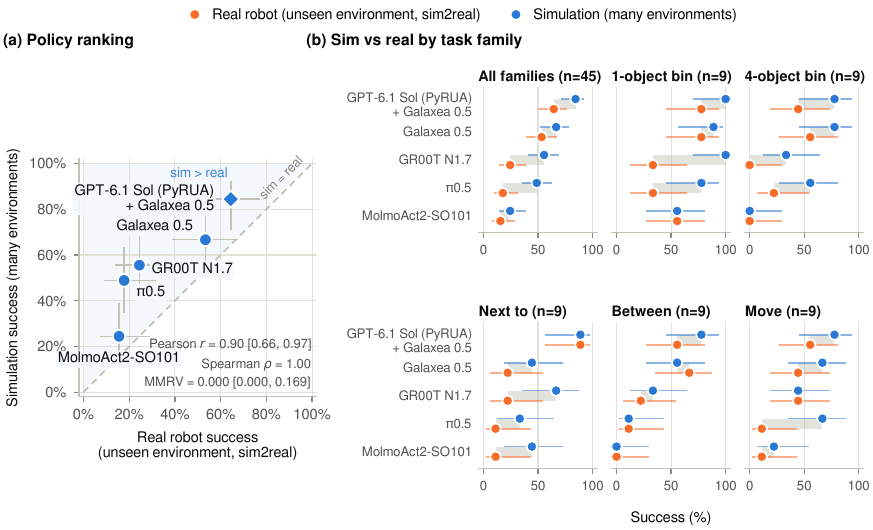}
  \par\vspace{5pt}
  \includegraphics[width=\textwidth]{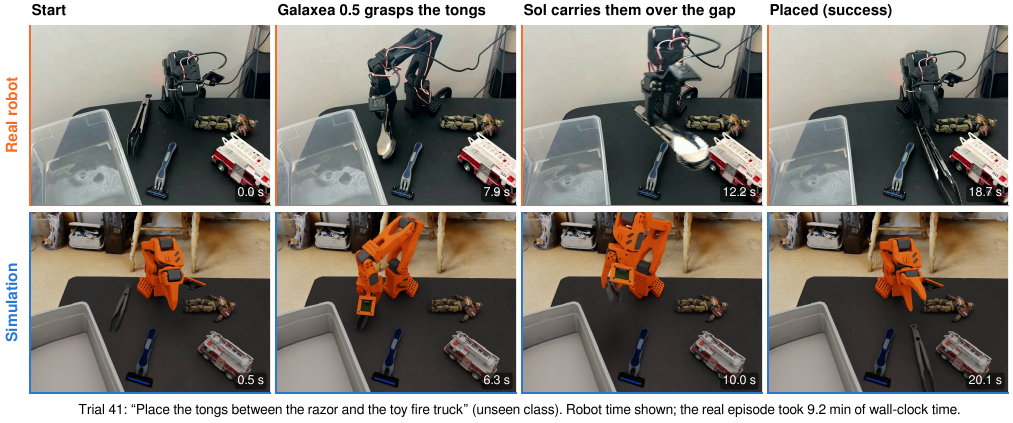}
  \caption{\textbf{Sim2real.} (a) Policy-level success on the real robot (unseen environment) vs.\ simulation
  (many environments) on the same 45 layouts: the ranking of all five policies is preserved exactly (Spearman
  $\rho=1.00$, MMRV $=0.000$) and every policy scores higher in simulation. Diamond: the LLM agent; circles: learned
  policies fine-tuned only on simulated demonstrations. (b) Success in simulation and on the real robot per
  policy, overall (45 episodes) and per task family (9 episodes), with 95\% Wilson intervals.
  Bottom: the agent on matched trial 41, a between episode with unseen-class objects that none of the four learned
  policies solved on the real robot. Galaxea 0.5 grasps the tongs, then the agent carries and places them, in both domains.}
  \label{fig:s2}
\end{figure}

A simulated benchmark is only useful if it tells us which policies are better in the real world. We test this
directly: five policies that never saw real data from our setup are evaluated, with identical weights, both in the
digital twin and on the physical SO-101.

\paragraph{Setup.} We evaluate four learned policies, Galaxea 0.5~\citep{liu2026g05}, GR00T~N1.7~\citep{nvidia2026gr00tn17},
$\pi_{0.5}$~\citep{pi2025pi05} and MolmoAct2-SO101~\citep{fang2026molmoact2}, and the
strongest LLM agent, GPT-6.1 Sol (PyRUA) + Galaxea 0.5, with the same checkpoints and harness as the leaderboard
(Sec.~\ref{sec:leaderboard}). The learned policies were fine-tuned only on simulated demonstrations; the agent pairs
a frozen language model with the same sim-trained Galaxea 0.5 checkpoint, to which it hands off grasping. The evaluation set
has 45 episodes: 3 for each of the 15 task-family $\times$ split cells, using only the 40 benchmark objects we still
own physically (18 seen, 11 unseen object, 11 unseen class), with every object used at least once. Each layout is
reproduced on the real table by projecting the simulated object footprints into the overhead camera view, so both
domains start from matched object positions and poses. The real robot operates in an environment that does not appear in
the training data, and real episodes are scored by the operator under the benchmark's success and failure rules
and the same time limits. As in simulation, the agent's time limit counts control steps only: the arm holds still
while the model computes. Median elapsed wall time to rollout stop is 197\,s for the agent, compared with
38\,s for Galaxea 0.5, 38\,s for GR00T~N1.7, 59\,s for $\pi_{0.5}$ and 56\,s for MolmoAct2-SO101 over the same
45 layouts, including successes and failures. These times come from timestamped control traces checked against
the original recordings and include inference waits and within-rollout stalls, but exclude model/server startup,
scene resets and subsequent operator scoring (Appendix~\ref{app:s2}). Lighting and
camera placement are fixed for the whole study. The four learned policies are evaluated in alternating blocks so
that slow drift does not favour any one of them, and the agent in a later session in the same environment; before
every episode the overhead view is compared with a reference frame (across all 225 real episodes the camera moved by
at most 2.6\,px and brightness changed by at most 7\%). In simulation each layout is run once with the canonical
appearance.

\paragraph{Simulation preserves the policy ranking.} Success in simulation and on the real robot ranks the five
policies identically (Fig.~\ref{fig:s2}a): GPT-6.1 Sol (PyRUA) + Galaxea 0.5 84\% (38/45) vs.\ 64\% (29/45), Galaxea 0.5 67\%
(30/45) vs.\ 53\% (24/45), GR00T~N1.7 56\% (25/45) vs.\ 24\% (11/45), $\pi_{0.5}$ 49\% (22/45) vs.\ 18\% (8/45), and
MolmoAct2-SO101 24\% (11/45) vs.\ 16\% (7/45). The Spearman correlation is $1.00$ and the mean maximum rank violation
(MMRV~\citep{li2024simpler}) is $0.000$ (95\% CI $[0.000, 0.169]$, resampling episodes within each policy); the
Pearson correlation is $0.90$ $[0.66, 0.97]$. The ranking spans two kinds of system, an LLM agent that plans and
places through code and four end-to-end policies, so it does not depend on the policies sharing an architecture.
With five policies these intervals remain wide, and we treat the result as evidence that simulation ranks policies
correctly rather than as a precise calibration.

\paragraph{The agent's lead transfers.} On the real robot the agent succeeds on 29 of 45 episodes against 24 for
Galaxea 0.5, the policy it calls for grasping: it solves 10 episodes that Galaxea 0.5 fails and fails 5 that Galaxea 0.5 solves (exact
McNemar $p=0.30$; in simulation 10 against 2, $p=0.04$). Its advantage lies where it, rather than Galaxea 0.5, does the
work. It succeeds on 8 of 9 real next-to episodes, against 2 for Galaxea 0.5, and on 7 of 15 unseen-class episodes, against
4. Eight real episodes were solved by the agent and by none of the four learned policies, including the between
episode with unseen-class objects in Fig.~\ref{fig:s2} (bottom). None of the agent's 16 real failures involved the
wrong object: 7 were on the bin tasks (5 knocked the lighter real bin), 3 moved a non-target object, and the
remaining 6 missed the placement criteria (for example, by setting the target on top of a referent).

\paragraph{Simulation is optimistic, by different amounts per policy.} Every policy succeeds more often in
simulation, by 9 to 31 percentage points (20 for the agent). Paired episodes agree on the outcome 56--91\% of the
time, and almost all disagreements are episodes that succeed in simulation and fail on the real robot (57, against
10 the other way).
The gap is largest on the bin tasks for GR00T~N1.7 and on the bin and move tasks for $\pi_{0.5}$
(Fig.~\ref{fig:s2}b): for example, GR00T bins the single object in 9/9 simulated episodes but 3/9 real ones; each policy's low
real-world success rate on bin tasks is, however, largely explained by the far lesser mass of the real plastic bin (making it easier to displace).
Many larger real objects were also scaled down to $0.7$--$0.9$ times their real-world size in simulation.
Overall, absolute success rates in simulation should be read as an upper bound on real-world performance, while comparisons between policies transfer. Per-family and
per-split results are in Appendix~\ref{app:s2}.


\section{Leaderboard}
\label{sec:leaderboard}

\begin{figure}[t]
  \centering
  \includegraphics[width=\textwidth]{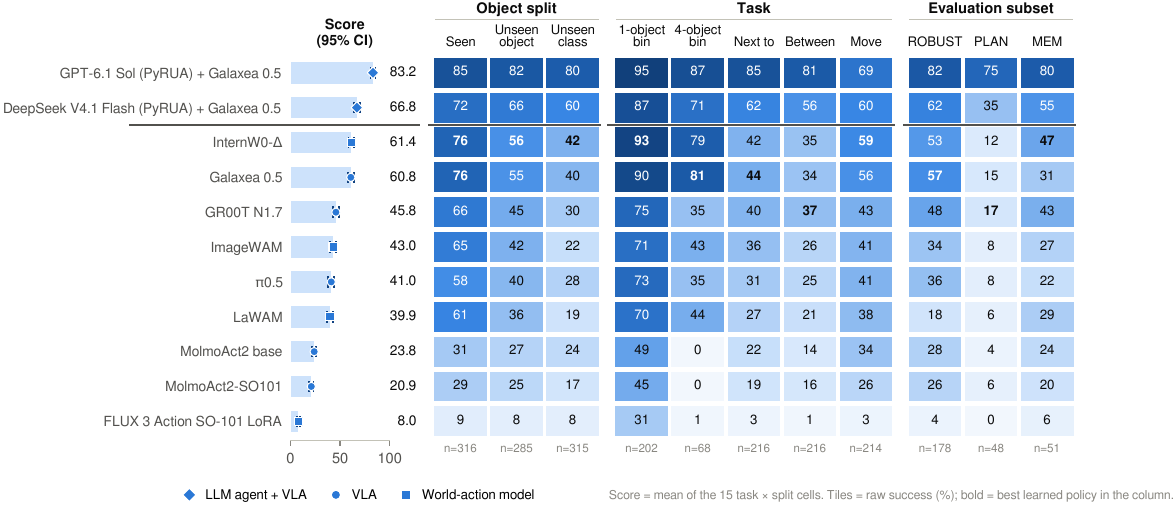}
  \caption{\textbf{SO-101 Bench leaderboard} (916 primary-evaluation episodes). Score is the mean success over the
  15 task $\times$ object-split cells with a 95\% bootstrap CI; tiles give raw success by object split, task and
  evaluation subset (ROBUST: held-out backgrounds and robot colours; PLAN: rotate to fit; MEM: remember occluded
  objects). Bold marks the best learned policy in each column. Diamonds: LLM agents; circles: VLAs; squares: WAMs.}
  \label{fig:s3}
\end{figure}

We evaluate nine learned policies and two LLM agents on the full primary evaluation (Fig.~\ref{fig:s3}).

\paragraph{Policies.} The learned policies cover two families. \emph{Vision-language-action models} (VLAs) map
images and language directly to actions: GR00T~N1.7~\citep{nvidia2026gr00tn17},
$\pi_{0.5}$~\citep{pi2025pi05}, Galaxea 0.5~\citep{liu2026g05}, and MolmoAct2~\citep{fang2026molmoact2}, which we fine-tune from
two starting points, the general MolmoAct2 checkpoint (\emph{base}) and its SO-100/101 post-trained variant
(\emph{SO101}). \emph{World-action models} (WAMs) additionally model future observations:
InternW0-$\Delta$~\citep{miao2026internw0} and FLUX~3 Action~\citep{bfl2026flux3action} jointly model video and actions,
ImageWAM~\citep{zhang2026imagewam} uses an image-editing backbone, and LaWAM~\citep{chen2026lawam}
predicts DINOv3~\citep{simeoni2025dinov3} latent states. 
All learned policies are fine-tuned on \texttt{so101\_bench\_sim} under the protocol of
Sec.~\ref{sec:protocol}; FLUX~3 Action is fine-tuned with LoRA~\citep{hu2021lora}, following its released recipe, and the rest are
fine-tuned in full or with their recommended frozen components (Appendix~\ref{app:s3}). The two \emph{LLM agents}
pair a frozen language model (GPT-6.1 Sol~\citep{openai2026sol} or DeepSeek V4.1 Flash-1017~\citep{deepseek2026flash})
with a code-as-policy harness~\citep{liang2023codeaspolicies}, PyRUA-Lean~\citep{si2026pyrua}, that
hands grasping to Galaxea 0.5; we analyse them in Sec.~\ref{sec:llm}.

\paragraph{Learned policies fall short of the benchmark.} The best learned policies, InternW0-$\Delta$
($61.4$, 95\% CI $[58.5, 64.1]$) and Galaxea 0.5 ($60.8$ $[57.9, 63.6]$), are statistically tied, followed by GR00T~N1.7
($45.8$), ImageWAM ($43.0$), $\pi_{0.5}$ ($41.0$) and LaWAM ($39.9$). Both MolmoAct2 variants score below 25 and
FLUX~3 Action scores 8.0. Starting MolmoAct2 from its SO-100/101 checkpoint does not help ($20.9$ vs.\ $23.8$ for
the base model, with overlapping intervals), so prior exposure to the embodiment (in its current form) is no substitute for the
benchmark's task and object coverage under this protocol.

\paragraph{Training scale appears to drive performance.} Under the equal 36\,hr budget, the number of fine-tuning
samples each policy sees varies by a factor of 30 (Appendix~\ref{app:s3}), and the amount of robot data in its
pretraining varies widely as well. The strongest learned policies combine large robot pretraining corpora with at
least a full epoch of fine-tuning (InternW0-$\Delta$ 1.2 epochs, Galaxea 0.5 1.6), whereas the three weakest saw well under
one epoch: MolmoAct2 0.5--0.6 epochs and FLUX~3 Action 0.2; both appear under-trained. The MolmoAct2 training
loss was still falling steeply when the budget ran out, and its published simulation recipe uses about five times
as many samples~\citep{fang2026molmoact2}; FLUX~3 Action's LoRA run used less than a tenth of the samples in its released full fine-tuning
recipe~\citep{bfl2026flux3action}. These results therefore rank policies at a fixed compute budget, and the low scores of MolmoAct2 and
FLUX~3 Action should not be read as their capability ceiling.

\paragraph{Spatial tasks and novel objects are the bottleneck.} Grasping familiar objects is largely solved: the
strongest learned policies bin a single object in 90--93\% of episodes and place 90\% of objects in 4-object bins.
The spatial tasks are far harder, with next-to and between success of 34--44\% for the same policies. Object
novelty compounds this. On the spatial tasks, InternW0-$\Delta$ falls from 72\% on seen objects to 18\% on unseen
classes for next to, and from 56\% to 19\% for between; Galaxea 0.5 falls from 71\% to 19\% and from 63\% to 12\%. Pooled
over tasks, the six strongest learned policies lose 30--43 points between seen objects and unseen classes; the
three weakest lose less only because their seen-object success is already low.

\paragraph{Subsets isolate distinct weaknesses.} PLAN is the clearest gap: no learned policy exceeds 17\% when the
grasped object must be rotated to fit (GR00T~N1.7 is best), while the agent reaches 75\%. On MEM,
learned policies reach at most 47\% (InternW0-$\Delta$, which has native memory~\citep{miao2026internw0}). ROBUST separates policies by visual robustness 
rather than skill: Galaxea 0.5 and GR00T~N1.7 keep their overall success under held-out backgrounds and robot colours, whereas LaWAM
drops from 39\% to 18\% and ImageWAM from 43\% to 34\%.

\paragraph{LLM agents lead.} Both agents outperform every learned policy, including the Galaxea 0.5 policy they call: GPT-6.1
Sol reaches $83.2$ $[80.7, 85.7]$ and DeepSeek V4.1 Flash-1017 $66.8$ $[63.7, 69.9]$. The agent's advantage is largest
where learned policies are weakest: Sol's success varies by only five points across object splits
(85/82/80\%) and it leads on PLAN and MEM. Sol also outperforms Galaxea 0.5 on the real robot (64\% vs.\ 53\%,
Sec.~\ref{sec:sim2real}). Section~\ref{sec:llm} analyses where this advantage comes from.


\begin{figure}[t]
  \centering
  \includegraphics[width=\textwidth]{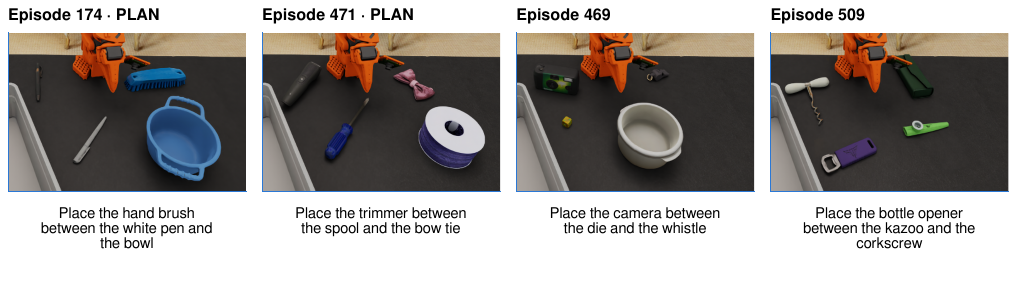}
  \caption{\textbf{Episodes solved only by GPT-6.1 Sol + PyRUA-Lean v9.6 + Galaxea 0.5.}
  Initial overhead views, cropped to the task workspace, with the exact instructions.
  All ten other leaderboard entries failed each episode. Episodes 174 and 471 belong to PLAN.}
  \label{fig:sol_only}
\end{figure}

\section{LLM Agents: Harness and Model Ablations}
\label{sec:llm}

The LLM agents lead the leaderboard (Sec.~\ref{sec:leaderboard}). We ask what drives this: the harness that
connects the language model to the robot, or the language model itself.

\paragraph{Harnesses.} All harnesses give the model the same overhead and wrist images, the same task rules and
no privileged simulator state. \emph{Direct tool calls} uses the \texttt{inspect-robots}
harness~\citep{robocurve2026inspectrobots}, letting the model command end-effector motions directly through tool calls. \emph{PyRUA-Lean}~\citep{si2026pyrua} is a code-as-policy harness~\citep{liang2023codeaspolicies}: the model writes Python that uses open-vocabulary
segmentation (via SAM3~\citep{carion2025sam3}), 3D localization and inverse kinematics to move the arm. \emph{PyRUA-Lean + Galaxea 0.5} adds the fine-tuned
Galaxea 0.5 policy as a tool: the model localizes objects, plans, and places, while handing grasping to Galaxea 0.5; the agent inspects Galaxea 0.5's
results periodically, and can initiate recoveries or take over. The simulator is paused while the model computes, so time limits apply to simulated time only.

\paragraph{Paired episodes.} Running agents on all 916 episodes is costly, so the ablations use 60 fixed
primary-evaluation episodes: 30 \emph{easy} episodes (six per task family) and 30 \emph{hard} episodes (six per
family with the fewest successes among the nine learned policies, including 11 PLAN episodes). Both sets were
selected before any agent was run.

\begin{figure}[t]
  \centering
  \includegraphics[width=\textwidth]{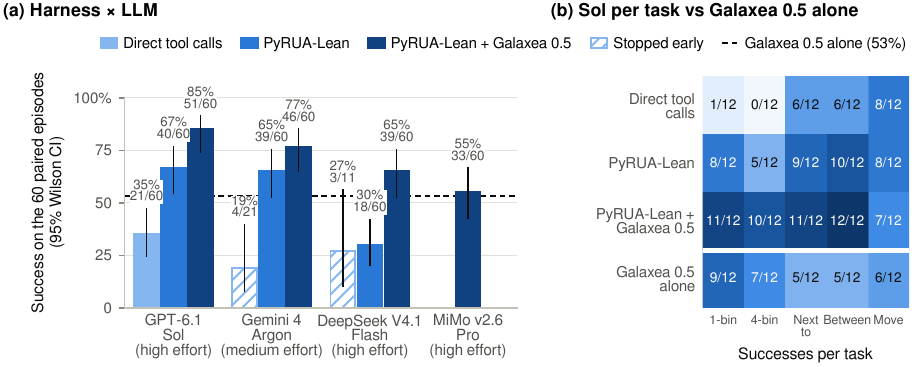}
  \caption{\textbf{Harness and LLM ablation} on the 60 paired episodes. (a) Success of each harness with three LLMs,
  and of the PyRUA-Lean v9.6 + Galaxea 0.5 harness with a fourth, MiMo v2.6 Pro (95\% Wilson intervals); hatched bars
  were stopped before completing all 60 episodes. The dashed line marks Galaxea 0.5 alone. (b) Successes per task family for GPT-6.1 Sol with each harness and for Galaxea 0.5 alone.}
  \label{fig:s4}
  \par\vspace{10pt}
  {\soInlineTableSetup
\begin{table}[t]
\centering
\small
\setlength{\tabcolsep}{3.5pt}
\caption{\textbf{Cost and token usage} of the PyRUA-Lean v9.6 + Galaxea 0.5 harness with each LLM, summed over the same 60 paired episodes. Output tokens include reasoning tokens (shown in parentheses). API cost is estimated from token counts at list prices (Gemini: gross Vertex rate).}
\label{tab:llm_cost}
\begin{tabular}{lcrrr}
\toprule
LLM (effort) & Success & API cost & Output tokens (reasoning) & Input tokens (cached) \\
\midrule
GPT-6.1 Sol (high) & 85\% (51/60) & \$14.68 & 0.32M (0.15M) & 61.82M (96\%) \\
Gemini 4 Argon (medium) & 77\% (46/60) & \$74.25 & 1.11M (0.86M) & 42.30M (73\%) \\
DeepSeek V4.1 Flash-1017 (high) & 65\% (39/60) & \$1.40 & 1.45M (1.23M) & 87.33M (98\%) \\
MiMo v2.6 Pro (high) & 55\% (33/60) & \$2.63 & 1.06M (0.82M) & 75.33M (96\%) \\
\midrule
Galaxea 0.5 alone & 53\% (32/60) & \$0 & -- & -- \\
\bottomrule
\end{tabular}
\end{table}
}
\end{figure}

\paragraph{The harness matters more than the model.} With GPT-6.1 Sol, success rises from 35\% (21/60) with
direct tool calls to 67\% (40/60) with PyRUA-Lean and 85\% (51/60) with the Galaxea 0.5 hand-off (Fig.~\ref{fig:s4}a),
against 53\% (32/60) for Galaxea 0.5 alone. Direct tool calls have trouble grasping (1/12 single-object bins,
0/12 four-object bins) but handle the spatial tasks reasonably (Fig.~\ref{fig:s4}b); writing perception and IK code
unlocks grasping, and handing grasps to Galaxea 0.5 makes them reliable. The hand-off is also cheaper: Sol's API cost
falls from \$18.44 with PyRUA-Lean to \$14.68 with the hand-off over the 60 episodes. The same ordering holds for Gemini~4 Argon (65\% to 77\%) and DeepSeek V4.1 Flash-1017 (30\% to 65\%).
The two policies are complementary: Galaxea 0.5 alone succeeds on 93\% of easy episodes but only 13\% of hard ones,
whereas Sol with the hand-off reaches 93\% and 77\%. Move is the one family where the hand-off does not help
(7/12, against 8/12 with PyRUA-Lean alone).

\paragraph{The model matters too.} Holding the PyRUA-Lean + Galaxea 0.5 harness fixed, we compare four language models
on the same 60 episodes (Fig.~\ref{fig:s4}a, Table~\ref{tab:llm_cost}). GPT-6.1 Sol~\citep{openai2026sol}
reaches 85\% (51/60), Gemini~4 Argon~\citep{google2026argon} 77\% (46/60),
DeepSeek V4.1 Flash-1017~\citep{deepseek2026flash} 65\% (39/60) and MiMo v2.6 Pro~\citep{xiaomi2026mimo} 55\% (33/60); all four exceed Galaxea 0.5 alone.
The models differ most on the spatial tasks (between: 4 to 12 of 12 episodes) and on the hard episodes, where
they solve 23, 20, 14 and 11 of 30. The full benchmark confirms the ordering of the two models we ran on all
916 episodes (83.2 vs.\ 66.8). Cost and capability are only loosely related: DeepSeek V4.1 Flash-1017 costs \$1.40 for the 60
episodes, a tenth of Sol, while Argon costs the most (\$74.25) without matching Sol. Sol also generates the fewest
tokens, 0.32M output tokens against 1.06--1.45M for the others (despite Gemini being at Medium effort).


\section{Conclusion}
\label{sec:conclusion}

SO-101 Bench makes evaluation of language-conditioned spatial manipulation on a low-cost arm both repeatable
and predictive of the real robot. Its digital twin reproduces which conditions are hard and why policies fail there,
and it ranks five policies trained without real data exactly as the physical SO-101 does; policies developed
against the benchmark run out of the box in unseen real environments with unseen objects. The leaderboard shows
where current methods stand: learned policies grasp familiar objects reliably but struggle with spatial placement,
unseen object classes and rotate-to-fit planning, while an LLM agent that writes code and delegates grasping to a
learned policy leads both in simulation and on the real robot, largely because of its harness.

\paragraph{Limitations.} The sim2real study covers five policies and 45 layouts in one real environment, so its
intervals are wide (MMRV 95\% CI $[0.000, 0.169]$), and simulation overestimates absolute success by 9 to 31
percentage points; simulated success rates should be read as an upper bound and comparisons, not levels, as what
transfers. The learned policies are compared at a fixed 36\,h fine-tuning budget, which left some of them
under-trained, so the leaderboard does not measure capability ceilings. The agents' time limits count simulated or
control time only: the simulator pauses, and the real arm holds still, while the language model computes, which
hides their real-time cost. Finally, the benchmark covers rigid objects and a single arm. Extending the twin to
deformable objects and bimanual setups, adding real-time constraints for agents, and using the twin's ranking as a
prior for cheaper real-robot evaluation are natural next steps.

A logical next step is to use repeated rollouts in the twin to refine control programs and accumulate validated,
reusable skills across episodes, as in ASPIRE~\citep{lu2026aspire} and RoboRSI~\citep{wen2026roborsi}, while
keeping the held-out evaluation fixed.


\clearpage

\bibliography{references}

\clearpage
\appendix

\section{GR00T Sim--Real Correspondence: Additional Results}
\label{app:s1}

\begin{table}[t]
\centering
\small
\caption{\textbf{GR00T sim--real correspondence.} A real-only GR00T N1.6-3B policy evaluated on the physical SO-101 versus an independently trained sim-only GR00T N1.6-3B policy evaluated in the digital twin, on 906 episodes matched on task, instruction, ordered objects and object split. $\Delta$ is sim minus real success in percentage points with a 95\% bootstrap CI that resamples matched semantic configurations within task$\times$split cells; agreement and MCC compare paired episode outcomes.}
\label{tab:corr_summary}
\begin{tabular}{lrrrrrr}
\toprule
& $n$ & Sim & Real & $\Delta$ (pp) [95\% CI] & Agree & MCC \\
\midrule
1-object bin & 368 & 73\% (268/368) & 86\% (316/368) & $-13.0$ $[-20.1, -6.5]$ & 81\% & $0.47$ \\
4-object bin & 56 & 18\% (10/56) & 46\% (26/56) & $-28.6$ $[-42.9, -14.3]$ & 64\% & $0.31$ \\
Next to & 178 & 25\% (45/178) & 48\% (85/178) & $-22.5$ $[-32.0, -12.4]$ & 49\% & $-0.04$ \\
Between & 187 & 21\% (40/187) & 37\% (69/187) & $-15.5$ $[-24.6, -6.4]$ & 60\% & $0.06$ \\
Move & 117 & 39\% (46/117) & 48\% (56/117) & $-8.5$ $[-20.5, +3.4]$ & 57\% & $0.14$ \\
\midrule
Seen & 397 & 59\% (236/397) & 71\% (281/397) & $-11.3$ $[-17.4, -5.5]$ & 67\% & $0.29$ \\
Unseen object, seen class & 316 & 41\% (131/316) & 62\% (195/316) & $-20.3$ $[-27.8, -13.2]$ & 62\% & $0.29$ \\
Unseen class & 193 & 22\% (42/193) & 39\% (76/193) & $-17.6$ $[-28.0, -8.3]$ & 72\% & $0.40$ \\
\midrule
\textbf{All} & 906 & 45\% (409/906) & 61\% (552/906) & $-15.8$ $[-19.9, -11.8]$ & 66\% & $0.36$ \\
\midrule
\multicolumn{7}{l}{Condition level (15 task$\times$split cells): Pearson $r=0.86$ [0.71, 0.91], Spearman $\rho=0.74$ [0.52, 0.91]} \\
\bottomrule
\end{tabular}
\end{table}

\begin{table}[t]
\centering
\small
\caption{\textbf{Failure-type correspondence (GR00T, next-to / between / move).} Every failed episode was annotated by hand in each domain and assigned to the first matching type in the order semantic $>$ grasping $>$ placement. Shares are of classified failures; TVD is the total variation distance between the real and sim mixes (0 = identical, 1 = disjoint).}
\label{tab:corr_failure_types}
\begin{tabular}{lrrrrrrrrr}
\toprule
& \multicolumn{4}{c}{Real} & \multicolumn{4}{c}{Sim} & \\
\cmidrule(lr){2-5}\cmidrule(lr){6-9}
& $n$ & Sem. & Grasp & Place & $n$ & Sem. & Grasp & Place & TVD \\
\midrule
All & 256 & 21\% & 24\% & 55\% & 286 & 24\% & 35\% & 41\% & 0.13 \\
\midrule
Seen & 97 & 9\% & 22\% & 69\% & 106 & 15\% & 23\% & 62\% & 0.07 \\
Unseen obj. / seen class & 102 & 24\% & 20\% & 57\% & 125 & 21\% & 41\% & 38\% & 0.21 \\
Unseen class & 57 & 37\% & 37\% & 26\% & 55 & 49\% & 44\% & 7\% & 0.19 \\
\midrule
\multicolumn{10}{l}{Episodes where both domains failed: 144; same type in 47\%, Cohen's $\kappa=0.17$} \\
\bottomrule
\end{tabular}
\end{table}

\begin{table}[t]
\centering
\small
\caption{\textbf{GR00T sim--real correspondence, all 15 task$\times$split cells.} Each entry is simulation / real success.}
\label{tab:corr_cells}
\begin{tabular}{lccc}
\toprule
Task & Seen & Unseen object, seen class & Unseen class \\
\midrule
1-object bin & 90\% (144/160) / 97\% (155/160) & 82\% (92/112) / 96\% (107/112) & 33\% (32/96) / 56\% (54/96) \\
4-object bin & 39\% (9/23) / 65\% (15/23) & 5\% (1/21) / 52\% (11/21) & 0\% (0/12) / 0\% (0/12) \\
Next to & 34\% (26/77) / 57\% (44/77) & 21\% (14/68) / 46\% (31/68) & 15\% (5/33) / 30\% (10/33) \\
Between & 34\% (29/85) / 45\% (38/85) & 11\% (8/76) / 38\% (29/76) & 12\% (3/26) / 8\% (2/26) \\
Move & 54\% (28/52) / 56\% (29/52) & 41\% (16/39) / 44\% (17/39) & 8\% (2/26) / 38\% (10/26) \\
\bottomrule
\end{tabular}
\end{table}

\begin{figure}[htbp]
  \centering
  \includegraphics[width=\textwidth]{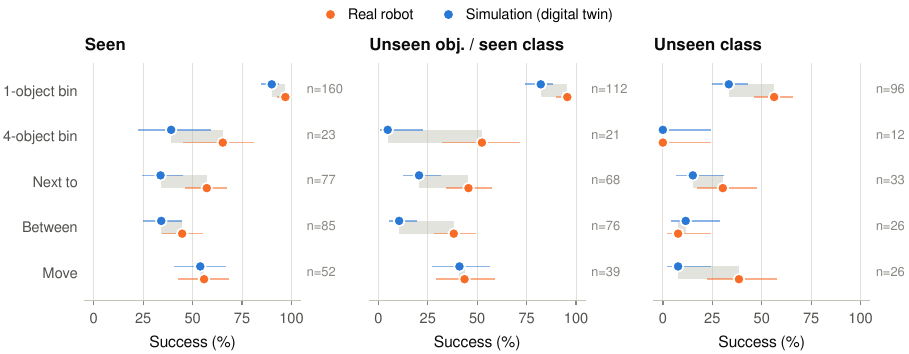}
  \caption{Real vs.\ simulation success in all 15 task $\times$ object-split cells (95\% Wilson intervals).}
  \label{fig:app_s1_cells}
\end{figure}

\begin{figure}[htbp]
  \centering
  \includegraphics[width=\textwidth]{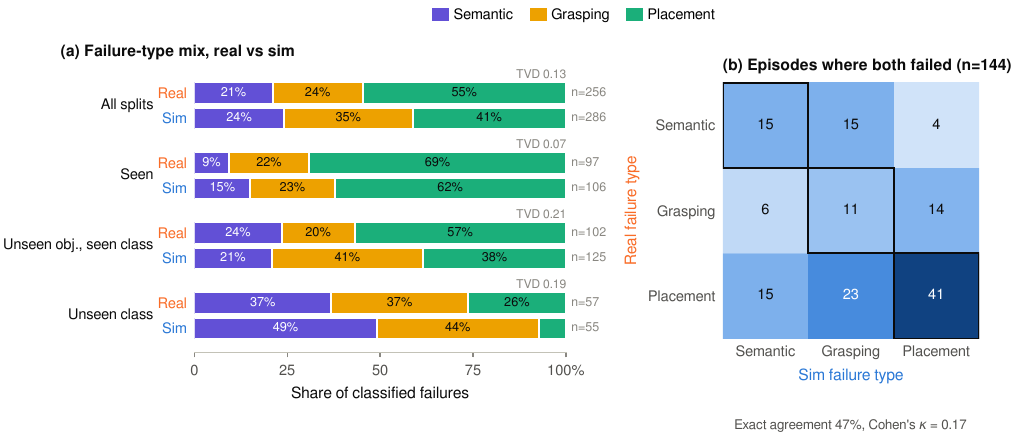}
  \caption{Failure-type mix per split and real $\times$ sim failure-type confusion on episodes where both domains
  failed.}
  \label{fig:app_s1_failtypes}
\end{figure}

\begin{figure}[htbp]
  \centering
  \includegraphics[width=\textwidth]{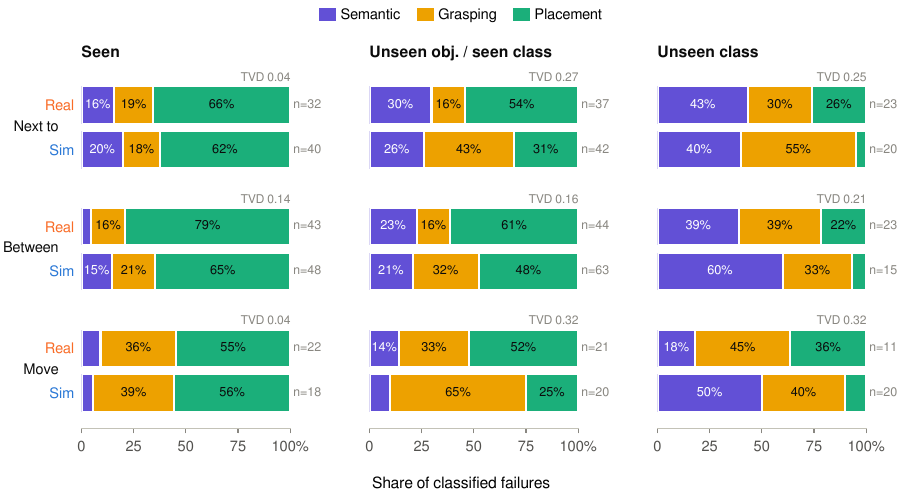}
  \caption{Failure-type mix per task and object split.}
  \label{fig:app_s1_failtypes_task}
\end{figure}

\begin{figure}[htbp]
  \centering
  \includegraphics[width=0.6\textwidth]{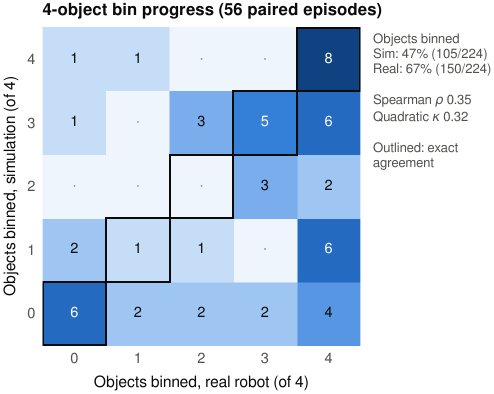}
  \caption{Objects binned in 4-object bin episodes, real vs.\ simulation (56 paired episodes).}
  \label{fig:app_s1_fourbin}
\end{figure}

\clearpage
\section{Sim2Real: Additional Results}
\label{app:s2}

\paragraph{The agent on the real robot.} GPT-6.1 Sol (PyRUA) + Galaxea 0.5 runs on the real robot with the same harness
(PyRUA-Lean v9.6), reasoning effort (high) and Galaxea 0.5 checkpoint as in simulation, with the real overhead and wrist
cameras in place of the simulated ones. Table~\ref{tab:real_wallclock} compares elapsed real-robot rollout time
for all five systems on the same 45 layouts. We use the last monotonic control timestamp relative to episode start
for every policy, checked against its original recording. Three older recordings omit elapsed gaps present in
the control traces, so compressed video duration is not substituted for elapsed time. These are deployed-system
measurements of time to rollout stop, not time to success; early failures can therefore shorten a rollout.

\begin{table}[htbp]
\centering
\small
\caption{\textbf{Real-robot elapsed time in Sim2Real.} Forty-five matched layouts per policy, including both successes and failures. Elapsed wall time runs from episode start to the last logged control command and includes policy/API waits and any within-rollout stalls; server/model startup, scene resets and subsequent operator scoring are excluded. Times measure rollout stopping, not time to successful completion.}
\label{tab:real_wallclock}
\begin{tabular}{lrrr}
\toprule
Policy & $n$ & Median (s) & P95 (s) \\
\midrule
GPT-6.1 Sol + PyRUA-Lean + Galaxea 0.5 & 45 & 196.6 & 391.8 \\
Galaxea 0.5 & 45 & 37.5 & 116.5 \\
GR00T N1.7 & 45 & 37.9 & 95.8 \\
$\pi_{0.5}$ & 45 & 59.1 & 153.3 \\
MolmoAct2-SO101 & 45 & 56.0 & 138.0 \\
\bottomrule
\end{tabular}
\end{table}

\begin{table}[t]
\centering
\small
\caption{\textbf{Sim2real evaluation with identical weights.} Each policy was evaluated on the same 45 layouts (5 task families $\times$ 3 object splits $\times$ 3) in simulation and on the physical SO-101. Agreement is the fraction of paired episodes with the same binary outcome. Ranking metrics are computed across the five policies; CIs resample episodes within each policy (10{,}000 replicates). MMRV~\citep{li2024simpler} is the mean maximum rank violation (0 = sim preserves the real ranking).}
\label{tab:mmrv_summary}
\begin{tabular}{lrrrr}
\toprule
Policy & Simulation & Real robot & $\Delta$ (pp) & Agreement \\
\midrule
GPT-6.1 Sol (PyRUA) + Galaxea 0.5 & 84\% (38/45) & 64\% (29/45) & $+20.0$ & 76\% (34/45) \\
Galaxea 0.5 & 67\% (30/45) & 53\% (24/45) & $+13.3$ & 64\% (29/45) \\
GR00T N1.7 & 56\% (25/45) & 24\% (11/45) & $+31.1$ & 56\% (25/45) \\
$\pi_{0.5}$ & 49\% (22/45) & 18\% (8/45) & $+31.1$ & 64\% (29/45) \\
MolmoAct2-SO101 & 24\% (11/45) & 16\% (7/45) & $+8.9$ & 91\% (41/45) \\
\midrule
& Pearson $r$ [95\% CI] & Spearman $\rho$ & MMRV [95\% CI] & \\
\cmidrule(lr){2-4}
Binary success & 0.90 [0.66, 0.97] & 1.00 & 0.000 [0.000, 0.169] & \\
4-bin as progress & 0.87 [0.62, 0.96] & 0.90 & 0.002 [0.000, 0.187] & \\
\bottomrule
\end{tabular}
\end{table}

\begin{figure}[htbp]
  \centering
  \includegraphics[width=0.8\textwidth]{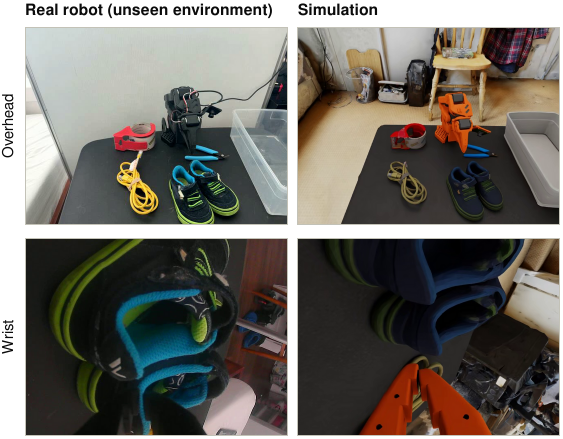}
  \caption{\textbf{One sim2real episode, real vs.\ simulation.} Initial overhead (top) and wrist (bottom) views of
  the same 4-object layout on the real robot and in the digital twin.}
  \label{fig:app_s2_example}
\end{figure}
\begin{table}[t]
\centering
\small
\caption{\textbf{Sim2real by task family and object split.} Successes in simulation / on the real robot; 9 episodes per family and 15 per split for each policy. GPT-6.1 Sol\textsuperscript{$\dagger$}: GPT-6.1 Sol (PyRUA) + Galaxea 0.5.}
\label{tab:mmrv_breakdown}
\begin{tabular}{lccccc}
\toprule
 & GPT-6.1 Sol\textsuperscript{$\dagger$} & Galaxea 0.5 & GR00T N1.7 & $\pi_{0.5}$ & MolmoAct2-SO101 \\
\midrule
1-object bin & 9 / 7 & 8 / 7 & 9 / 3 & 7 / 3 & 5 / 5 \\
4-object bin & 7 / 4 & 7 / 5 & 3 / 0 & 5 / 2 & 0 / 0 \\
Next to & 8 / 8 & 4 / 2 & 6 / 2 & 3 / 1 & 4 / 1 \\
Between & 7 / 5 & 5 / 6 & 3 / 2 & 1 / 1 & 0 / 0 \\
Move & 7 / 5 & 6 / 4 & 4 / 4 & 6 / 1 & 2 / 1 \\
\midrule
Seen & 15 / 12 & 14 / 11 & 10 / 5 & 12 / 5 & 5 / 3 \\
Unseen object, seen class & 13 / 10 & 10 / 9 & 10 / 4 & 7 / 2 & 5 / 3 \\
Unseen class & 10 / 7 & 6 / 4 & 5 / 2 & 3 / 1 & 1 / 1 \\
\bottomrule
\end{tabular}
\end{table}

\begin{figure}[htbp]
  \centering
  \includegraphics[width=0.75\textwidth]{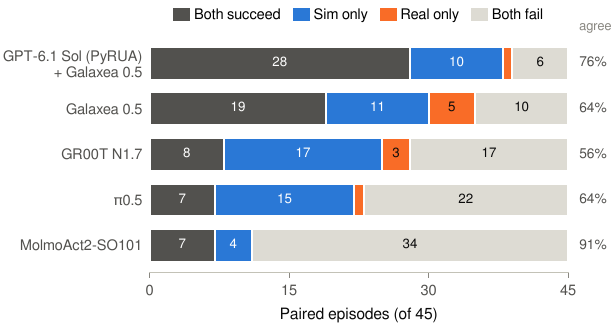}
  \caption{Paired episode outcomes per policy.}
  \label{fig:app_s2_agreement}
\end{figure}

\begin{figure}[htbp]
  \centering
  \includegraphics[width=\textwidth]{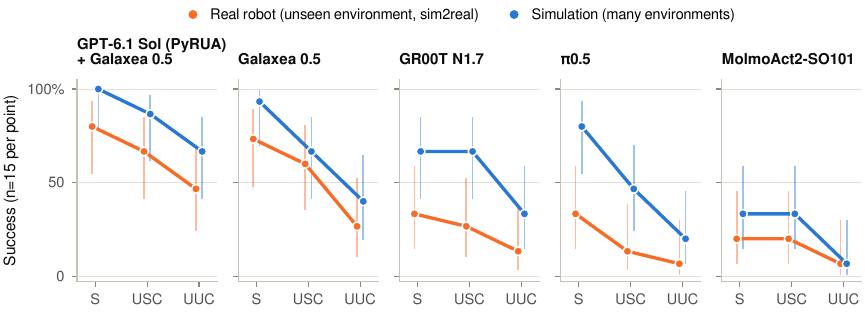}
  \caption{Seen $\to$ unseen gradient in each domain, per policy.}
  \label{fig:app_s2_splits}
\end{figure}

\clearpage
\section{Leaderboard: Additional Results}
\label{app:s3}

\paragraph{Measured inference latency.} Table~\ref{tab:policy_latency} summarizes server-reported
\texttt{dt\_ms} values from the completed primary-evaluation runs, after removing the same eight excluded trials
as the leaderboard. Each sample times one action-chunk query, including preprocessing and action conversion
within the server's prediction call; it excludes HTTP transport, simulation and model loading. PNG decoding is
inside some server timers and outside others. All recorded first calls are retained. ImageWAM uses precomputed
instruction embeddings. The shared HTTP client labels these messages ``MolmoAct2 inference'' for all eight
HTTP-served policies; policy identity comes from the completed run and checkpoint, not that generic log label.
GR00T~N1.7 did not record this metric and is marked NR. The returned chunk length can exceed the prefix actually
executed before replanning, so chunk latency is not per-control-step latency or an end-to-end control rate.

\begin{table}[htbp]
\centering
\small
\caption{\textbf{Measured learned-policy inference latency.} Server-reported wall time per action-chunk query over the 916 scored primary-evaluation episodes, pooled across completed runs (one evaluation worker). First calls are retained. Steps are returned / executed, not necessarily the underlying model's full prediction horizon. Median and 95th percentile (P95) are in milliseconds. Hardware differs across policies; these are deployment measurements, not a controlled hardware comparison. NR: not recorded.}
\label{tab:policy_latency}
\begin{tabular}{llrrrr}
\toprule
Policy & GPU & Steps & Queries & Median (ms) & P95 (ms) \\
\midrule
GR00T N1.7 & RTX 3090 & -- / 16 & NR & NR & NR \\
$\pi_{0.5}$ & RTX A5000 & 30 / 15 & 44,334 & 337.0 & 376.9 \\
Galaxea 0.5 & RTX A5000 & 27 / 15 & 33,422 & 791.8 & 933.8 \\
MolmoAct2-SO101 & RTX A5000 & 30 / 30 & 26,046 & 696.8 & 741.3 \\
MolmoAct2 base & RTX A5000 & 30 / 30 & 25,661 & 775.3 & 818.6 \\
FLUX 3 Action SO-101 LoRA & A40 & 42 / 32 & 25,080 & 2894.0 & 3129.4 \\
ImageWAM & RTX A5000 & 16 / 16 & 41,721 & 322.5 & 410.0 \\
LaWAM & RTX A5000 & 36 / 36 & 18,651 & 274.1 & 333.4 \\
InternW0-$\Delta$ & A40 & 16 / 16 & 34,378 & 986.5 & 1089.8 \\
\bottomrule
\end{tabular}
\end{table}

\begin{table*}[t]
\centering
\small
\setlength{\tabcolsep}{4pt}
\caption{\textbf{SO-101 Bench leaderboard} (primary evaluation, 916 episodes). The score is the mean success over the 15 task $\times$ object-split cells, with a 95\% bootstrap CI that resamples episodes within cells. Remaining columns are raw success rates (\%) over the $n$ episodes in the header. All learned policies were fine-tuned on \texttt{so101\_bench\_sim}; LLM agents are frozen PyRUA-Lean v9.6 harnesses that hand grasping to Galaxea 0.5 and perform placement themselves.}
\label{tab:leaderboard}
\begin{tabular}{rllcrrrrrrrrr}
\toprule
& Policy & Type & Score [95\% CI] & 1-bin & 4-bin & Next to & Between & Move & S & USC & UUC & 4-bin prog. \\
& & & & \scriptsize $n$=202 & \scriptsize $n$=68 & \scriptsize $n$=216 & \scriptsize $n$=216 & \scriptsize $n$=214 & \scriptsize 316 & \scriptsize 285 & \scriptsize 315 & \\
\midrule
1 & GPT-6.1 Sol (PyRUA) + Galaxea 0.5 & LLM agent + VLA & \textbf{83.2} [80.7, 85.7] & 95 & 87 & 85 & 81 & 69 & 85 & 82 & 80 & 94 \\
2 & DeepSeek V4.1 Flash-1017 (PyRUA) + Galaxea 0.5 & LLM agent + VLA & \textbf{66.8} [63.7, 69.9] & 87 & 71 & 62 & 56 & 60 & 72 & 66 & 60 & 81 \\
\midrule
3 & InternW0-$\Delta$ & WAM & \textbf{61.4} [58.5, 64.1] & 93 & 79 & 42 & 35 & 59 & 76 & 56 & 42 & 91 \\
4 & Galaxea 0.5 & VLA & \textbf{60.8} [57.9, 63.6] & 90 & 81 & 44 & 34 & 56 & 76 & 55 & 40 & 90 \\
5 & GR00T N1.7 & VLA & \textbf{45.8} [42.6, 48.8] & 75 & 35 & 40 & 37 & 43 & 66 & 45 & 30 & 52 \\
6 & ImageWAM & WAM & \textbf{43.0} [40.2, 45.7] & 71 & 43 & 36 & 26 & 41 & 65 & 42 & 22 & 58 \\
7 & $\pi_{0.5}$ & VLA & \textbf{41.0} [37.8, 44.3] & 73 & 35 & 31 & 25 & 41 & 58 & 40 & 28 & 65 \\
8 & LaWAM & WAM & \textbf{39.9} [36.8, 42.9] & 70 & 44 & 27 & 21 & 38 & 61 & 36 & 19 & 60 \\
9 & MolmoAct2 base & VLA & \textbf{23.8} [21.5, 26.1] & 49 & 0 & 22 & 14 & 34 & 31 & 27 & 24 & 18 \\
10 & MolmoAct2-SO101 & VLA & \textbf{20.9} [18.6, 23.2] & 45 & 0 & 19 & 16 & 26 & 29 & 25 & 17 & 22 \\
11 & FLUX 3 Action SO-101 LoRA & WAM & \textbf{8.0} [6.5, 9.5] & 31 & 1 & 3 & 1 & 3 & 9 & 8 & 8 & 15 \\
\bottomrule
\end{tabular}
\end{table*}

\begin{table*}[t]
\centering
\scriptsize
\setlength{\tabcolsep}{3pt}
\caption{\textbf{Leaderboard, all 15 task $\times$ object-split cells.} Success as \% (successes/episodes).}
\label{tab:leaderboard_cells}
\begin{tabular}{lccccccccccccccc}
\toprule
 & \multicolumn{3}{c}{1-object bin} & \multicolumn{3}{c}{4-object bin} & \multicolumn{3}{c}{Next to} & \multicolumn{3}{c}{Between} & \multicolumn{3}{c}{Move} \\
\cmidrule(lr){2-4}\cmidrule(lr){5-7}\cmidrule(lr){8-10}\cmidrule(lr){11-13}\cmidrule(lr){14-16}
Policy & S & USC & UUC & S & USC & UUC & S & USC & UUC & S & USC & UUC & S & USC & UUC \\
\midrule
GPT-6.1 Sol (PyRUA) + Galaxea 0.5 & \makecell{99\%\\(66/67)} & \makecell{92\%\\(59/64)} & \makecell{93\%\\(66/71)} & \makecell{96\%\\(23/24)} & \makecell{77\%\\(17/22)} & \makecell{86\%\\(19/22)} & \makecell{85\%\\(64/75)} & \makecell{91\%\\(61/67)} & \makecell{80\%\\(59/74)} & \makecell{84\%\\(63/75)} & \makecell{82\%\\(55/67)} & \makecell{77\%\\(57/74)} & \makecell{72\%\\(54/75)} & \makecell{65\%\\(42/65)} & \makecell{69\%\\(51/74)} \\
DeepSeek V4.1 Flash-1017 (PyRUA) + Galaxea 0.5 & \makecell{97\%\\(65/67)} & \makecell{88\%\\(56/64)} & \makecell{76\%\\(54/71)} & \makecell{88\%\\(21/24)} & \makecell{77\%\\(17/22)} & \makecell{45\%\\(10/22)} & \makecell{67\%\\(50/75)} & \makecell{64\%\\(43/67)} & \makecell{54\%\\(40/74)} & \makecell{60\%\\(45/75)} & \makecell{49\%\\(33/67)} & \makecell{57\%\\(42/74)} & \makecell{61\%\\(46/75)} & \makecell{60\%\\(39/65)} & \makecell{59\%\\(44/74)} \\
InternW0-$\Delta$ & \makecell{99\%\\(66/67)} & \makecell{98\%\\(63/64)} & \makecell{82\%\\(58/71)} & \makecell{92\%\\(22/24)} & \makecell{86\%\\(19/22)} & \makecell{59\%\\(13/22)} & \makecell{72\%\\(54/75)} & \makecell{34\%\\(23/67)} & \makecell{18\%\\(13/74)} & \makecell{56\%\\(42/75)} & \makecell{28\%\\(19/67)} & \makecell{19\%\\(14/74)} & \makecell{76\%\\(57/75)} & \makecell{55\%\\(36/65)} & \makecell{46\%\\(34/74)} \\
Galaxea 0.5 & \makecell{97\%\\(65/67)} & \makecell{92\%\\(59/64)} & \makecell{82\%\\(58/71)} & \makecell{92\%\\(22/24)} & \makecell{86\%\\(19/22)} & \makecell{64\%\\(14/22)} & \makecell{71\%\\(53/75)} & \makecell{40\%\\(27/67)} & \makecell{19\%\\(14/74)} & \makecell{63\%\\(47/75)} & \makecell{27\%\\(18/67)} & \makecell{12\%\\(9/74)} & \makecell{72\%\\(54/75)} & \makecell{54\%\\(35/65)} & \makecell{42\%\\(31/74)} \\
GR00T N1.7 & \makecell{94\%\\(63/67)} & \makecell{77\%\\(49/64)} & \makecell{55\%\\(39/71)} & \makecell{67\%\\(16/24)} & \makecell{23\%\\(5/22)} & \makecell{14\%\\(3/22)} & \makecell{61\%\\(46/75)} & \makecell{43\%\\(29/67)} & \makecell{16\%\\(12/74)} & \makecell{56\%\\(42/75)} & \makecell{25\%\\(17/67)} & \makecell{27\%\\(20/74)} & \makecell{57\%\\(43/75)} & \makecell{42\%\\(27/65)} & \makecell{30\%\\(22/74)} \\
ImageWAM & \makecell{93\%\\(62/67)} & \makecell{77\%\\(49/64)} & \makecell{46\%\\(33/71)} & \makecell{88\%\\(21/24)} & \makecell{36\%\\(8/22)} & \makecell{0\%\\(0/22)} & \makecell{56\%\\(42/75)} & \makecell{36\%\\(24/67)} & \makecell{15\%\\(11/74)} & \makecell{51\%\\(38/75)} & \makecell{18\%\\(12/67)} & \makecell{8\%\\(6/74)} & \makecell{56\%\\(42/75)} & \makecell{42\%\\(27/65)} & \makecell{24\%\\(18/74)} \\
$\pi_{0.5}$ & \makecell{85\%\\(57/67)} & \makecell{72\%\\(46/64)} & \makecell{63\%\\(45/71)} & \makecell{58\%\\(14/24)} & \makecell{27\%\\(6/22)} & \makecell{18\%\\(4/22)} & \makecell{51\%\\(38/75)} & \makecell{31\%\\(21/67)} & \makecell{12\%\\(9/74)} & \makecell{43\%\\(32/75)} & \makecell{21\%\\(14/67)} & \makecell{11\%\\(8/74)} & \makecell{55\%\\(41/75)} & \makecell{40\%\\(26/65)} & \makecell{28\%\\(21/74)} \\
LaWAM & \makecell{87\%\\(58/67)} & \makecell{69\%\\(44/64)} & \makecell{55\%\\(39/71)} & \makecell{67\%\\(16/24)} & \makecell{45\%\\(10/22)} & \makecell{18\%\\(4/22)} & \makecell{52\%\\(39/75)} & \makecell{27\%\\(18/67)} & \makecell{3\%\\(2/74)} & \makecell{51\%\\(38/75)} & \makecell{9\%\\(6/67)} & \makecell{3\%\\(2/74)} & \makecell{56\%\\(42/75)} & \makecell{40\%\\(26/65)} & \makecell{18\%\\(13/74)} \\
MolmoAct2 base & \makecell{57\%\\(38/67)} & \makecell{45\%\\(29/64)} & \makecell{45\%\\(32/71)} & \makecell{0\%\\(0/24)} & \makecell{0\%\\(0/22)} & \makecell{0\%\\(0/22)} & \makecell{19\%\\(14/75)} & \makecell{25\%\\(17/67)} & \makecell{22\%\\(16/74)} & \makecell{27\%\\(20/75)} & \makecell{10\%\\(7/67)} & \makecell{5\%\\(4/74)} & \makecell{36\%\\(27/75)} & \makecell{35\%\\(23/65)} & \makecell{31\%\\(23/74)} \\
MolmoAct2-SO101 & \makecell{46\%\\(31/67)} & \makecell{44\%\\(28/64)} & \makecell{44\%\\(31/71)} & \makecell{0\%\\(0/24)} & \makecell{0\%\\(0/22)} & \makecell{0\%\\(0/22)} & \makecell{27\%\\(20/75)} & \makecell{18\%\\(12/67)} & \makecell{11\%\\(8/74)} & \makecell{25\%\\(19/75)} & \makecell{16\%\\(11/67)} & \makecell{5\%\\(4/74)} & \makecell{31\%\\(23/75)} & \makecell{31\%\\(20/65)} & \makecell{16\%\\(12/74)} \\
FLUX 3 Action SO-101 LoRA & \makecell{31\%\\(21/67)} & \makecell{31\%\\(20/64)} & \makecell{31\%\\(22/71)} & \makecell{4\%\\(1/24)} & \makecell{0\%\\(0/22)} & \makecell{0\%\\(0/22)} & \makecell{7\%\\(5/75)} & \makecell{1\%\\(1/67)} & \makecell{0\%\\(0/74)} & \makecell{1\%\\(1/75)} & \makecell{0\%\\(0/67)} & \makecell{3\%\\(2/74)} & \makecell{3\%\\(2/75)} & \makecell{5\%\\(3/65)} & \makecell{3\%\\(2/74)} \\
\bottomrule
\end{tabular}
\end{table*}

\begin{table*}[t]
\centering
\small
\caption{\textbf{Evaluation subsets and visual conditions.} ROBUST (held-out backgrounds and robot colours), PLAN and MEM are tagged subsets of the same 916 episodes; appearance profiles partition them. Profiles use different layouts, so the columns are not a paired visual-robustness test.}
\label{tab:leaderboard_subsets}
\begin{tabular}{lcccccc}
\toprule
& \multicolumn{3}{c}{Subsets} & \multicolumn{3}{c}{Appearance profile} \\
\cmidrule(lr){2-4}\cmidrule(lr){5-7}
Policy & ROBUST & PLAN & MEM & Canonical & Varied & Held-out \\
\midrule
GPT-6.1 Sol (PyRUA) + Galaxea 0.5 & 82\% (146/178) & 75\% (36/48) & 80\% (41/51) & 82\% (307/374) & 83\% (303/364) & 82\% (146/178) \\
DeepSeek V4.1 Flash-1017 (PyRUA) + Galaxea 0.5 & 62\% (111/178) & 35\% (17/48) & 55\% (28/51) & 66\% (248/374) & 68\% (246/364) & 62\% (111/178) \\
InternW0-$\Delta$ & 53\% (94/178) & 12\% (6/48) & 47\% (24/51) & 59\% (219/374) & 60\% (220/364) & 53\% (94/178) \\
Galaxea 0.5 & 57\% (102/178) & 15\% (7/48) & 31\% (16/51) & 57\% (214/374) & 57\% (209/364) & 57\% (102/178) \\
GR00T N1.7 & 48\% (85/178) & 17\% (8/48) & 43\% (22/51) & 49\% (184/374) & 45\% (164/364) & 48\% (85/178) \\
ImageWAM & 34\% (60/178) & 8\% (4/48) & 27\% (14/51) & 47\% (174/374) & 44\% (159/364) & 34\% (60/178) \\
$\pi_{0.5}$ & 36\% (64/178) & 8\% (4/48) & 22\% (11/51) & 45\% (167/374) & 41\% (151/364) & 36\% (64/178) \\
LaWAM & 18\% (32/178) & 6\% (3/48) & 29\% (15/51) & 46\% (173/374) & 42\% (152/364) & 18\% (32/178) \\
MolmoAct2 base & 28\% (50/178) & 4\% (2/48) & 24\% (12/51) & 30\% (112/374) & 24\% (88/364) & 28\% (50/178) \\
MolmoAct2-SO101 & 26\% (47/178) & 6\% (3/48) & 20\% (10/51) & 23\% (86/374) & 24\% (86/364) & 26\% (47/178) \\
FLUX 3 Action SO-101 LoRA & 4\% (7/178) & 0\% (0/48) & 6\% (3/51) & 10\% (39/374) & 9\% (34/364) & 4\% (7/178) \\
\bottomrule
\end{tabular}
\end{table*}

\begin{table}[t]
\centering
\scriptsize
\caption{\textbf{Evaluated checkpoints.} Hugging Face repository and revision (learned policies) or frozen harness (LLM agents), and the evaluation hardware.}
\label{tab:leaderboard_checkpoints}
\begin{tabular}{lll}
\toprule
Policy & Checkpoint & Hardware \\
\midrule
GPT-6.1 Sol (PyRUA) + Galaxea 0.5 & \texttt{PyRUA-Lean v9.6 + G0.5 step25600 (@\allowbreak 957d33b)} & RunPod A40, 2 isolated agent workers per pod \\
DeepSeek V4.1 Flash-1017 (PyRUA) + Galaxea 0.5 & \texttt{PyRUA-Lean v9.6 + G0.5 step25600 (@\allowbreak 957d33b)} & RunPod A40, 2 isolated agent workers per pod \\
InternW0-$\Delta$ & \texttt{5hadytru/so101\_bench\_InternW0-Delta@\allowbreak 8316ccc} & RunPod A40, 1 worker \\
Galaxea 0.5 & \texttt{5hadytru/so101\_bench\_G0.5\_full\_fine\_tune@\allowbreak 957d33b (step25600)} & RunPod RTX A5000 secure, 1 worker \\
GR00T N1.7 & \texttt{5hadytru/so101\_bench\_GR00T\_N1.7-3B@\allowbreak 572a932 (step16000)} & workstation RTX 3090, 1 worker \\
ImageWAM & \texttt{5hadytru/so101\_bench\_ImageWAM-FLUX.2-4B\_full\_fine\_tune@\allowbreak 0ba49ae} & RunPod RTX A5000, 1 worker \\
$\pi_{0.5}$ & \texttt{5hadytru/so101\_bench\_pi05@\allowbreak de712e6 (step18000)} & RunPod RTX A5000 secure, 1 worker \\
LaWAM & \texttt{5hadytru/so101\_bench\_LaWAM\_full\_fine\_tune@\allowbreak a7ddd36} & RunPod RTX A5000, 1 worker \\
MolmoAct2 base & \texttt{5hadytru/so101\_bench\_MolmoAct2-SO101\_WM\_base\_full\_fine\_tune@\allowbreak 723a9a8} & RunPod RTX A5000, 1 worker \\
MolmoAct2-SO101 & \texttt{5hadytru/so101\_bench\_MolmoAct2-SO101\_WM\_full\_fine\_tune@\allowbreak 8d04677} & RunPod RTX A5000, 1 worker \\
FLUX 3 Action SO-101 LoRA & \texttt{5hadytru/so101\_bench\_flux-3-action-so101@\allowbreak ac49157} & RunPod A40, 1 worker \\
\bottomrule
\end{tabular}
\end{table}

\begin{sidewaystable*}
\centering
\scriptsize
\setlength{\tabcolsep}{3pt}
\caption{\textbf{Full SO-101 Bench leaderboard.} All entries are evaluated on the same 916 episodes of the primary evaluation. Score = mean success over the 15 task $\times$ object-split cells, with a 95\% bootstrap CI (episodes resampled within cells). S / USC / UUC = seen / unseen object, seen class / unseen class. 4-bin prog.\ = mean fraction of objects binned in 4-object bin episodes. Sim time = total simulated time. GPU-h for learned policies = sim time $\times$ measured wall/sim ratio on the listed GPU; agent GPU-hours are not reported. API cost is a token-price estimate. Samples = updates $\times$ global batch; epochs over the 2.0M training frames; \% of LIBERO recipe = samples relative to that model's official LIBERO fine-tune. A100-h = 4 GPUs $\times$ hours to the evaluated checkpoint (estimated from measured s/update). so101\_bench\_sim\_WM adds episode-start (init) frames to so101\_bench\_sim's overhead and wrist cameras. ``?'' = not yet recorded.}
\label{tab:full_leaderboard}
\begin{tabular}{llrrrrrrrrrrrrrr}
\toprule
\multicolumn{16}{l}{\textbf{(a) Performance} (success \%, $n$ episodes in each header)} \\
\midrule
Policy & Type & Score [95\% CI] & All & S & USC & UUC & 1-bin & 4-bin & Next to & Between & Move & 4-bin prog. & MEM & PLAN & ROBUST \\
& & & 916 & 316 & 285 & 315 & 202 & 68 & 216 & 216 & 214 & 68 & 51 & 48 & 178 \\
\midrule
GPT-6.1 Sol (PyRUA) + Galaxea 0.5 & LLM agent + VLA & 83.2 [80.7, 85.7] & 83 & 85 & 82 & 80 & 95 & 87 & 85 & 81 & 69 & 94 & 80 & 75 & 82 \\
DeepSeek V4.1 Flash-1017 (PyRUA) + Galaxea 0.5 & LLM agent + VLA & 66.8 [63.7, 69.9] & 66 & 72 & 66 & 60 & 87 & 71 & 62 & 56 & 60 & 81 & 55 & 35 & 62 \\
InternW0-Delta & WAM (video generation) & 61.4 [58.5, 64.1] & 58 & 76 & 56 & 42 & 93 & 79 & 42 & 35 & 59 & 91 & 47 & 12 & 53 \\
Galaxea 0.5 & VLA & 60.8 [57.9, 63.6] & 57 & 76 & 55 & 40 & 90 & 81 & 44 & 34 & 56 & 90 & 31 & 15 & 57 \\
GR00T N1.7 & VLA & 45.8 [42.6, 48.8] & 47 & 66 & 45 & 30 & 75 & 35 & 40 & 37 & 43 & 52 & 43 & 17 & 48 \\
ImageWAM & WAM (image editing) & 43.0 [40.2, 45.7] & 43 & 65 & 42 & 22 & 71 & 43 & 36 & 26 & 41 & 58 & 27 & 8 & 34 \\
$\pi$0.5 & VLA & 41.0 [37.8, 44.3] & 42 & 58 & 40 & 28 & 73 & 35 & 31 & 25 & 41 & 65 & 22 & 8 & 36 \\
LaWAM & WAM (latent action) & 39.9 [36.8, 42.9] & 39 & 61 & 36 & 19 & 70 & 44 & 27 & 21 & 38 & 60 & 29 & 6 & 18 \\
MolmoAct2 base & VLA & 23.8 [21.5, 26.1] & 27 & 31 & 27 & 24 & 49 & 0 & 22 & 14 & 34 & 18 & 24 & 4 & 28 \\
MolmoAct2-SO101 & VLA & 20.9 [18.6, 23.2] & 24 & 29 & 25 & 17 & 45 & 0 & 19 & 16 & 26 & 22 & 20 & 6 & 26 \\
FLUX 3 Action SO-101 LoRA & WAM (video generation) & 8.0 [6.5, 9.5] & 9 & 9 & 8 & 8 & 31 & 1 & 3 & 1 & 3 & 15 & 6 & 0 & 4 \\
\bottomrule
\end{tabular}
\vspace{6pt}
\begin{tabular}{llrrrr}
\toprule
\multicolumn{6}{l}{\textbf{(b) Evaluation cost} (916 episodes)} \\
\midrule
Policy & Hardware & Sim time (h) & GPU-h & API \$ & API \$/episode \\
\midrule
GPT-6.1 Sol (PyRUA) + Galaxea 0.5 & A40, 2 agent workers per GPU & 5.7 & -- & 216 & 0.236 \\
DeepSeek V4.1 Flash-1017 (PyRUA) + Galaxea 0.5 & A40, 2 agent workers per GPU & 5.8 & -- & 22 & 0.024 \\
InternW0-Delta & A40 & 5.4 & 50 & -- & -- \\
Galaxea 0.5 & RTX A5000 & 5.0 & 44 & -- & -- \\
GR00T N1.7 & RTX 3090 & 6.6 & 39 & -- & -- \\
ImageWAM & RTX A5000 & 6.5 & 53 & -- & -- \\
$\pi$0.5 & RTX A5000 & 6.5 & 54 & -- & -- \\
LaWAM & RTX A5000 & 6.5 & 59 & -- & -- \\
MolmoAct2 base & RTX A5000 & 7.4 & 59 & -- & -- \\
MolmoAct2-SO101 & RTX A5000 & 7.5 & 56 & -- & -- \\
FLUX 3 Action SO-101 LoRA & A40 & 7.7 & 105 & -- & -- \\
\bottomrule
\end{tabular}
\vspace{6pt}
\begin{tabular}{p{3.2cm}p{4.2cm}p{3.6cm}rrrrrl}
\toprule
\multicolumn{9}{l}{\textbf{(c) Fine-tuning} (to the evaluated checkpoint; equal 36 h budget on 4$\times$A100-80GB)} \\
\midrule
Policy & Trainable parameters & Memory & Updates $\times$ batch & Samples & Epochs & \% of LIBERO recipe & A100-h & W\&B \\
\midrule
GPT-6.1 Sol (PyRUA) + Galaxea 0.5 & 0 (LLM); G0.5 as below & Agent context (full episode) + G0.5's frame history & \multicolumn{6}{l}{LLM frozen (API); uses the G0.5 checkpoint below} \\
DeepSeek V4.1 Flash-1017 (PyRUA) + Galaxea 0.5 & 0 (LLM); G0.5 as below & Agent context (full episode) + G0.5's frame history & \multicolumn{6}{l}{LLM frozen (API); uses the G0.5 checkpoint below} \\
InternW0-Delta & Full (Wan2.2-5B video expert + 1.2B action expert; RynnBrain-2B VLM frozen) & Episode-start anchor frame + frame 32 steps back & 18{,}200 $\times$ 128 & 2.33M & 1.17 & ? & 142 & \href{https://wandb.ai/5hadytru/finetune-internw0-so101/runs/12stqtva}{run} \\
Galaxea 0.5 & Full (8-bit AdamW); count TBD & 6 observation frames at 1 s stride & 25{,}600 $\times$ 128 & 3.28M & 1.64 & ? & 134 & \href{https://wandb.ai/5hadytru/finetune-g05-so101/runs/vpuf2wct}{run} \\
GR00T N1.7 & Projector + diffusion action head + VLLN (LLM and vision frozen) & Episode-start frames (overhead + wrist init) & 16{,}000 $\times$ 640 & 10.24M & 5.13 & 80\% & 142 & \href{https://wandb.ai/5hadytru/finetune-gr00t-n1d7}{project} \\
ImageWAM & FLUX.2-klein-4B editing branch + $\sim$642M action DiT (text encoder frozen) & None (current frame) & 30{,}000 $\times$ 128 & 3.84M & 1.92 & ? & 130 & \href{https://wandb.ai/5hadytru/finetune-imagewam-so101/runs/y0hgogns}{run} \\
$\pi$0.5 & Full, 4.14B & Episode-start frames (overhead + wrist init) & 9{,}000 $\times$ 256 & 2.30M & 1.15 & 30\% & 136 & \href{https://wandb.ai/5hadytru/finetune-pi05-so101/runs/g260uy3k}{run} \\
LaWAM & Full (Qwen3-VL-2B + DINOv3 + LAM) & None (current frame) & 45{,}000 $\times$ 256 & 11.52M & 5.77 & ? & 132 & \href{https://wandb.ai/5hadytru/finetune-lawam-so101/runs/s4obq0gm}{run} \\
MolmoAct2 base & Full, $\sim$5.4B & Episode-start frames (overhead + wrist init) & 8{,}000 $\times$ 128 & 1.02M & 0.51 & 16\% & 110 & \href{https://wandb.ai/5hadytru/finetune-molmoact2-so101/runs/76ech8au}{run} \\
MolmoAct2-SO101 & Full, $\sim$5.4B & Episode-start frames (overhead + wrist init) & 10{,}000 $\times$ 128 & 1.28M & 0.64 & 20\% & 129 & \href{https://wandb.ai/5hadytru/finetune-molmoact2-so101/runs/krwz749d}{run} \\
FLUX 3 Action SO-101 LoRA & LoRA r32 on attention/MLP + action heads (full) & 8 ticks (0.27 s) of state history + 2 frames & 3{,}000 $\times$ 128 & 0.38M & 0.19 & ? & 69 & \href{https://wandb.ai/5hadytru/finetune-flux3-so101/runs/9ixare4e}{run} \\
\bottomrule
\end{tabular}
\vspace{6pt}
\begin{tabular}{p{3.2cm}p{4.6cm}p{3.0cm}p{3.6cm}p{3.6cm}}
\toprule
\multicolumn{5}{l}{\textbf{(d) Data}} \\
\midrule
Policy & Initial checkpoint & SO-101 in pre-training & SO-101 in post-training & Fine-tuning dataset \\
\midrule
GPT-6.1 Sol (PyRUA) + Galaxea 0.5 & gpt-6.1-sol (API, frozen) + G0.5 step 25600 & via G0.5 & via G0.5 & none (LLM frozen); G0.5 as below \\
DeepSeek V4.1 Flash-1017 (PyRUA) + Galaxea 0.5 & deepseek-flash-1017 (API, frozen) + G0.5 step 25600 & via G0.5 & via G0.5 & none (LLM frozen); G0.5 as below \\
InternW0-Delta & InternRobotics/InternW0-Delta-Base @ ca534d4 & ? & No & so101\_bench\_sim \\
Galaxea 0.5 & Galaxea g05-base & ? & ? & so101\_bench\_sim \\
GR00T N1.7 & nvidia/GR00T-N1.7-3B & ? & No & so101\_bench\_sim\_WM (+ init frames) \\
ImageWAM & yuyangalin/ImageWAM-FLUX.2-4B-InternData-A1-EE @ 51a71ed & No & No & so101\_bench\_sim \\
$\pi$0.5 & lerobot/pi05\_base @ b211f3d & ? & No & so101\_bench\_sim\_WM (+ init frames) \\
LaWAM & jialei02/lawam-pretrain-lerobot @ bc45090 & No & No & so101\_bench\_sim \\
MolmoAct2 base & allenai/MolmoAct2 @ e432d85 & Yes (post-training mix includes SO-100/101) & Yes (in multi-embodiment mix) & so101\_bench\_sim\_WM (+ init frames) \\
MolmoAct2-SO101 & MolmoAct2-SO100\_101 (MolmoAct2 post-trained 100k updates on SO-100/101) & Yes & Yes (dedicated SO-100/101 checkpoint) & so101\_bench\_sim\_WM (+ init frames) \\
FLUX 3 Action SO-101 LoRA & flux-3-action-so101 (BFL) & ? & Yes (1,016 community SO-100/101 datasets) & so101\_bench\_sim \\
\bottomrule
\end{tabular}
\end{sidewaystable*}

\begin{figure}[htbp]
  \centering
  \includegraphics[width=\textwidth]{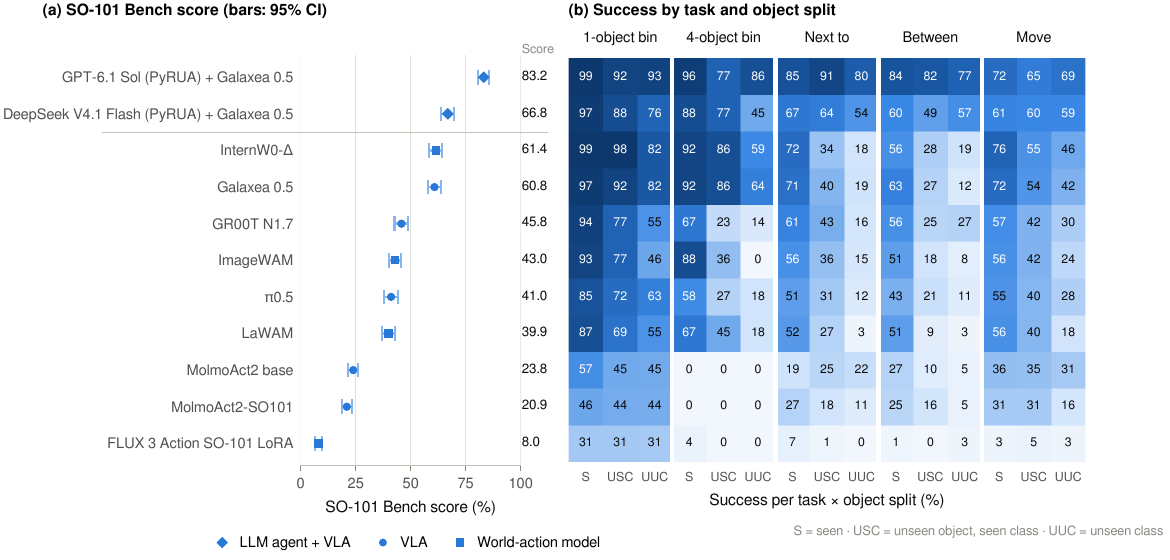}
  \caption{Score and success in all 15 task $\times$ object-split cells.}
  \label{fig:app_s3_cells}
\end{figure}

\begin{figure}[htbp]
  \centering
  \includegraphics[width=0.8\textwidth]{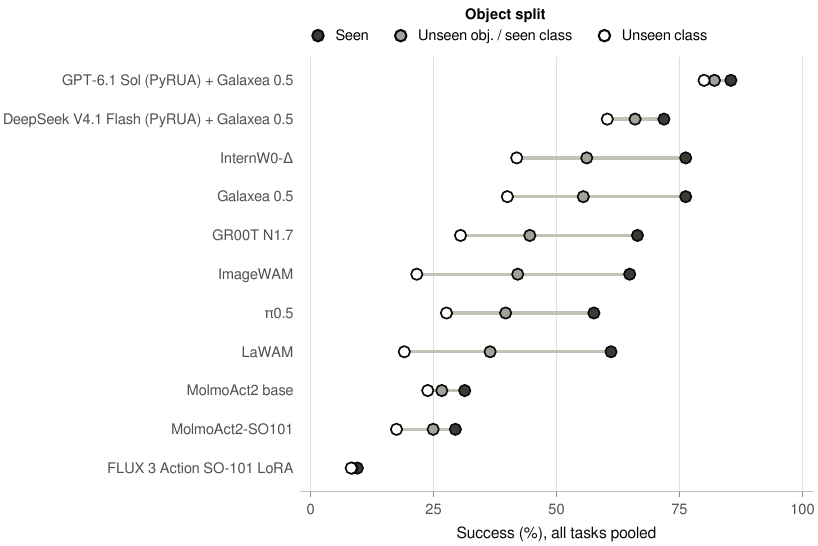}
  \caption{Success by object split, all tasks pooled.}
  \label{fig:app_s3_splits}
\end{figure}

\begin{figure}[htbp]
  \centering
  \includegraphics[width=0.8\textwidth]{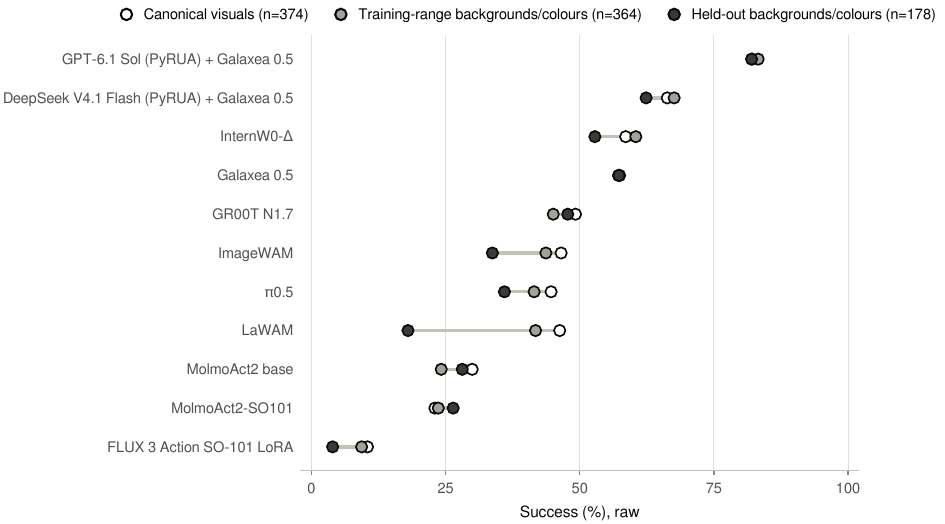}
  \caption{Success by appearance profile: canonical, training-range and held-out visuals.}
  \label{fig:app_s3_appearance}
\end{figure}

\begin{figure}[htbp]
  \centering
  \includegraphics[width=\textwidth]{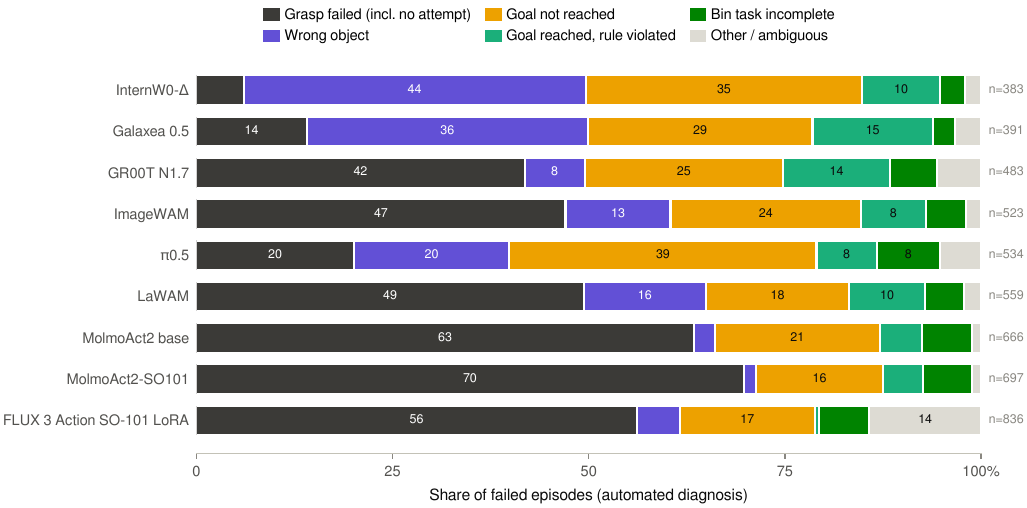}
  \caption{Automated failure diagnosis for the learned policies. \todo{Only include after validating the
  classifier against human labels.}}
  \label{fig:app_s3_failures}
\end{figure}

\clearpage
\section{LLM Agents: Additional Results}
\label{app:llm}

\begin{table*}[t]
\centering
\small
\setlength{\tabcolsep}{4pt}
\caption{\textbf{Harness ablation} on 60 paired primary-evaluation episodes (30 easy, 30 hard; six per task in each stratum, selected from the nine learned policies' outcomes before any LLM run). Direct: the LLM issues end-effector tool calls. PyRUA-Lean: code-as-policy with perception and IK tools. + Galaxea 0.5: the same harness hands grasping to Galaxea 0.5; the LLM performs placement. API cost is an estimate at list prices. $^\dagger$Stopped by the operator after the listed episodes (easy stratum only).}
\label{tab:harness}
\begin{tabular}{llcccccccc}
\toprule
LLM & Harness & All 60 & Easy 30 & Hard 30 & 1-bin & 4-bin & Next to & Between & Move \\
\midrule
GPT-6.1 Sol & Direct tool calls & 35\% (21/60) & 47\% (14/30) & 23\% (7/30) & 1/12 & 0/12 & 6/12 & 6/12 & 8/12 \\
 & PyRUA-Lean & 67\% (40/60) & 73\% (22/30) & 60\% (18/30) & 8/12 & 5/12 & 9/12 & 10/12 & 8/12 \\
 & PyRUA-Lean + Galaxea 0.5 & 85\% (51/60) & 93\% (28/30) & 77\% (23/30) & 11/12 & 10/12 & 11/12 & 12/12 & 7/12 \\
\midrule
Gemini 4 Argon & Direct tool calls$^\dagger$ & 19\% (4/21) & 19\% (4/21) & -- & 0/5 & 0/5 & 0/4 & 0/3 & 4/4 \\
 & PyRUA-Lean & 65\% (39/60) & 70\% (21/30) & 60\% (18/30) & 8/12 & 2/12 & 9/12 & 12/12 & 8/12 \\
 & PyRUA-Lean + Galaxea 0.5 & 77\% (46/60) & 87\% (26/30) & 67\% (20/30) & 9/12 & 9/12 & 9/12 & 10/12 & 9/12 \\
\midrule
DeepSeek V4.1 Flash-1017 & Direct tool calls$^\dagger$ & 27\% (3/11) & 27\% (3/11) & -- & 0/2 & 0/3 & 0/2 & 1/2 & 2/2 \\
 & PyRUA-Lean & 30\% (18/60) & 43\% (13/30) & 17\% (5/30) & 4/12 & 0/12 & 5/12 & 6/12 & 3/12 \\
 & PyRUA-Lean + Galaxea 0.5 & 65\% (39/60) & 83\% (25/30) & 47\% (14/30) & 12/12 & 9/12 & 6/12 & 4/12 & 8/12 \\
\midrule
\multicolumn{2}{l}{Galaxea 0.5 alone (reference)} & 53\% (32/60) & 93\% (28/30) & 13\% (4/30) & 9/12 & 7/12 & 5/12 & 5/12 & 6/12 \\
\bottomrule
\end{tabular}
\end{table*}

\begin{table*}[t]
\centering
\small
\setlength{\tabcolsep}{4pt}
\caption{\textbf{LLM ablation} with the PyRUA-Lean v9.6 + Galaxea 0.5 harness held fixed, on the same 60 paired episodes. Effort is the provider reasoning setting. API cost is an estimate at list prices (Gemini: gross Vertex list rate). The last column repeats the full-benchmark score where the hybrid was run on all 916 episodes.}
\label{tab:llm}
\begin{tabular}{lllccccccccr}
\toprule
LLM & Effort & Runtime & All 60 & Easy & Hard & 1-bin & 4-bin & Next to & Between & Move & \$/episode \\
\midrule
GPT-6.1 Sol & high & Codex & 85\% (51/60) & 93\% (28/30) & 77\% (23/30) & 11/12 & 10/12 & 11/12 & 12/12 & 7/12 & 0.245 \\
Gemini 4 Argon & medium & OpenAI-compatible chat (Vertex) & 77\% (46/60) & 87\% (26/30) & 67\% (20/30) & 9/12 & 9/12 & 9/12 & 10/12 & 9/12 & 1.237 \\
DeepSeek V4.1 Flash-1017 & high & Codex & 65\% (39/60) & 83\% (25/30) & 47\% (14/30) & 12/12 & 9/12 & 6/12 & 4/12 & 8/12 & 0.023 \\
MiMo v2.6 Pro & high & Codex via Responses$\to$Chat bridge & 55\% (33/60) & 73\% (22/30) & 37\% (11/30) & 7/12 & 8/12 & 7/12 & 7/12 & 4/12 & 0.044 \\
\midrule
Galaxea 0.5 alone & -- & -- & 53\% (32/60) & 93\% (28/30) & 13\% (4/30) & 9/12 & 7/12 & 5/12 & 5/12 & 6/12 & 0 \\
\bottomrule
\end{tabular}
\end{table*}

\begin{figure}[htbp]
  \centering
  \includegraphics[width=\textwidth]{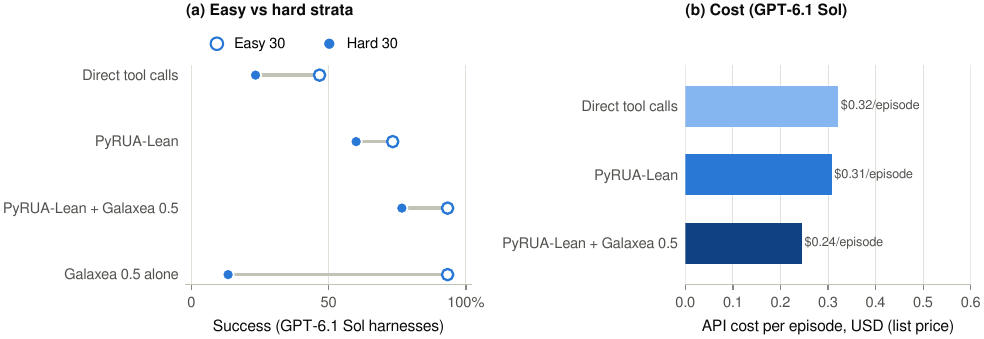}
  \caption{Harness ablation by stratum (easy30 / hard30) and cost.}
  \label{fig:app_s4_strata}
\end{figure}

\begin{figure}[htbp]
  \centering
  \includegraphics[width=\textwidth]{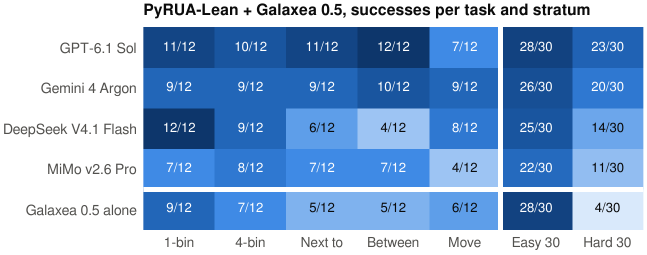}
  \caption{Successes per task and stratum for each LLM, and for Galaxea 0.5 alone.}
  \label{fig:app_s5_task}
\end{figure}

\clearpage
\section{Object Real2Sim Pipeline}
\label{app:objects}

Every object passes through the same pipeline, which produces a rigid body with separate visual and collision
geometry.

\paragraph{1. Asset generation.} We start from an image of the object: a photograph of the real object, an image
from the internet, or an image from a text-to-image model. An image-to-3D model accessed through 3D AI Studio~\citep{3daistudio2026} converts it into a textured mesh
(GLB) with physically based rendering (PBR) materials.

\paragraph{2. Mesh processing (Blender).} The mesh is scaled to the object's real dimensions, oriented, and given
an origin at its base, which serves as the object's root pose for the success conditions. Texture colours are
corrected towards the object's real palette and brightness. The mesh is then duplicated into a \emph{visual} and a
\emph{physics} mesh.
\begin{itemize}
  \item \textbf{Visual mesh:} remeshed and decimated to a fixed visual budget, repaired (loose geometry removed,
  duplicate vertices merged, no non-manifold edges), and the original colour and roughness are baked onto it.
  \item \textbf{Physics mesh:} cleaned and remeshed to a coarser physical budget, avoiding small crevices in which
  the gripper could become caught. Over-faceted results are smoothed, remaining non-manifold regions are filled or
  remeshed at a larger voxel size, and face normals are made consistent. The physics mesh is checked to align with
  the visual mesh.
\end{itemize}
Both meshes are exported together as a USD file with packed textures.

\paragraph{3. Physics setup (Isaac Sim).} The root prim is made a rigid body with mass. The physics mesh is
given a collider, using a signed-distance-field (SDF) collider for complex non-convex shapes without thin walls
and a convex hull or convex decomposition otherwise. A physics material sets static and dynamic friction and
restitution, and the physics mesh is hidden from rendering.

\paragraph{4. Appearance tuning (Isaac Lab).} The visual mesh's colour texture is adjusted, usually by tuning its
brightness and occasionally by repainting regions, until the object's appearance under the benchmark cameras
matches the real object.

\clearpage
\section{Simulation Backgrounds}
\label{app:backgrounds}

\begin{figure}[p]
  \centering
  \includegraphics[width=\textwidth,height=0.9\textheight,keepaspectratio]{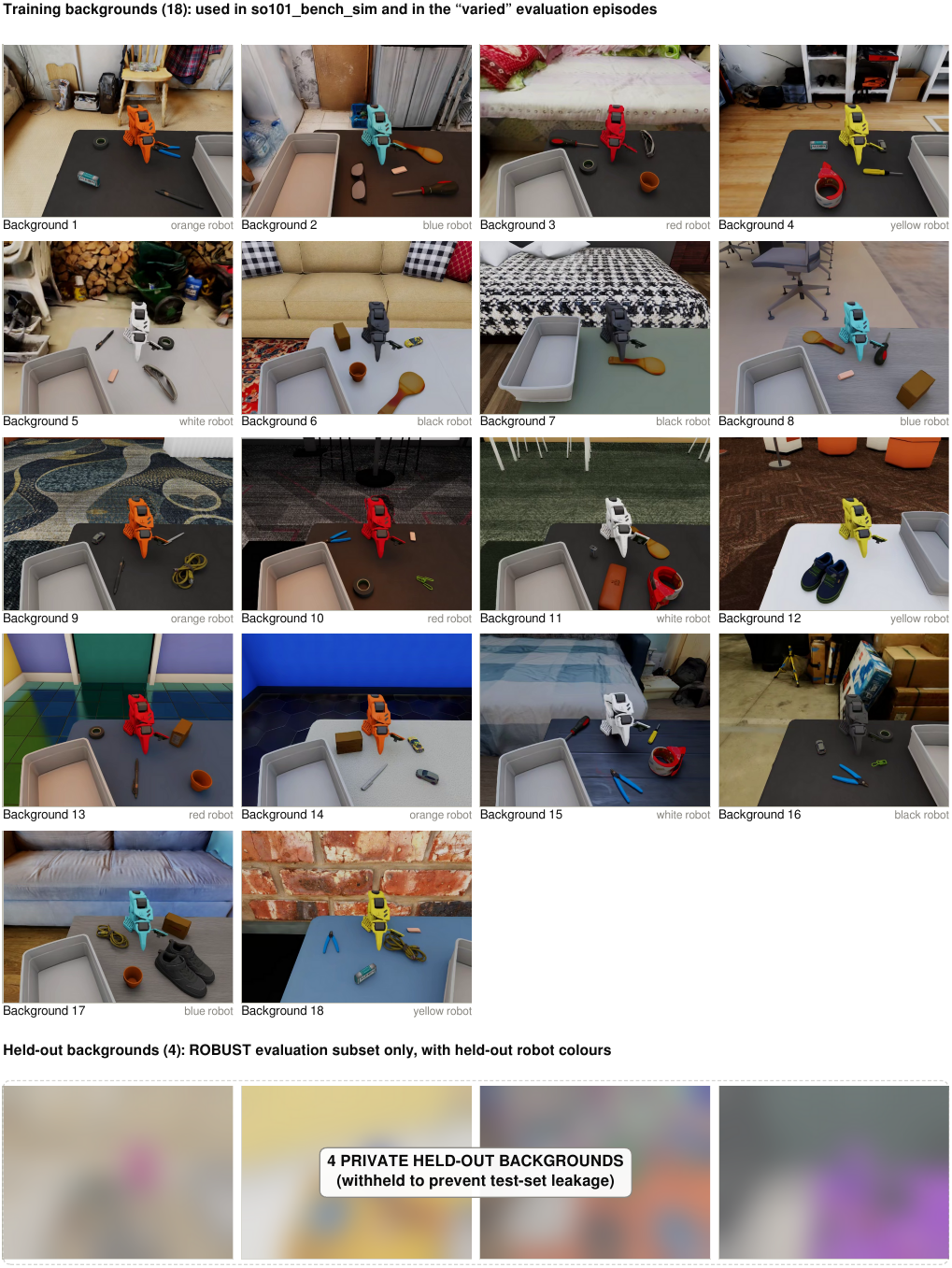}
  \caption{\textbf{Simulation backgrounds.} The 18 training backgrounds (shown with varied robot colours) and the
  4 held-out ROBUST backgrounds, which are withheld to prevent test-set leakage.}
  \label{fig:backgrounds}
\end{figure}

\end{document}